\documentclass[authoryear,final,5p,times,twocolumn]{elsarticle}

\usepackage{graphicx}
\usepackage{amsmath,amsfonts}
\usepackage{amssymb}
\usepackage{algorithm}
\usepackage{algpseudocode}
\usepackage{booktabs}
\usepackage{multirow}
\usepackage{placeins}
\usepackage{float}
\usepackage{subcaption}
\usepackage{adjustbox}
\usepackage[hidelinks]{hyperref}

\journal{Expert Systems with Applications}

\begin{document}
\raggedbottom

\begin{frontmatter}

\title{HDMamba-YOLO: Efficient State-Space Perception and Local Spatial Reconstruction for UAV Small Object Detection}

\author[aff1]{Linduo Wei}
\ead{weilinduo@njust.edu.cn}

\author[aff2]{Junjie Fan}
\ead{junjiefan@njust.edu.cn}

\author[aff3]{Yijun Mai}
\ead{maiyijun@njust.edu.cn}

\author[aff3]{Xiao Chen}
\ead{xiaochen25@njust.edu.cn}

\author[aff1]{Guiyang Zhang}

\author[aff2]{Yong Qi\corref{cor1}}
\ead{qyong@njust.edu.cn}
\cortext[cor1]{Corresponding author: Yong Qi.}

\affiliation[aff1]{
    organization={School of Economics and Management, Nanjing University of Science and Technology},
    city={Nanjing},
    postcode={210094},
    country={China}
}

\affiliation[aff2]{
    organization={School of Intellectual Property, Nanjing University of Science and Technology},
    city={Nanjing},
    postcode={210094},
    country={China}
}

\affiliation[aff3]{
    organization={School of Computer Science and Engineering, Nanjing University of Science and Technology},
    city={Nanjing},
    postcode={210094},
    country={China}
}

\begin{abstract}
Small objects in UAV imagery provide limited appearance and boundary cues, making them difficult to distinguish from cluttered backgrounds and localize accurately. A central challenge is to incorporate long-range context efficiently while retaining the weak spatial evidence needed for precise localization throughout multi-scale feature aggregation. To address this challenge, we propose Hybrid Dual-domain Mamba-YOLO (HDMamba-YOLO), a heterogeneous state-space--convolutional detector that assigns complementary modeling roles to different detection stages. The backbone uses efficient state-space modeling to capture long-range contextual dependencies, while the neck performs local two-dimensional feature reconstruction through convolutional processing. At the core of this reconstruction, Adaptive Spatial-Semantic Attention Fusion (ASSAF) adaptively combines local-detail and broader neighborhood responses during feature aggregation. Content-adaptive resampling reduces spatial mismatch across scales, while macro--micro feature interaction provides task-specific modulation for classification and localization. On VisDrone2019, HDMamba-YOLO-B achieves 42.737\% mAP$_{50}$ and 25.713\% mAP$_{50:95}$ with 10.042M parameters and 29.879 corrected GFLOPs. Under the unified AI-TOD evaluation protocol, it achieves 21.621\% AP and 47.881\% AP$_{50}$. Controlled ablations support the stage-wise allocation of contextual perception and local reconstruction and indicate complementary contributions from adaptive alignment and task-specific interaction in UAV small-object detection.
\end{abstract}


\end{frontmatter}

\section{Introduction}

UAV-based visual sensing has become an important tool for traffic monitoring, emergency response, and other large-area perception tasks \citep{obaid2025uav,wang2025lsod}. Its flexible aerial viewpoint enables rapid acquisition of high-resolution scenes, but also produces a detection setting dominated by small and densely distributed objects. In low-altitude UAV imagery, many targets occupy only 8--32 pixels and exhibit weak textures and boundaries against heterogeneous backgrounds \citep{du2019visdrone,cao2021visdrone,wang2021tiny,nikouei2025small}. Their limited visual evidence increases the risk of missed detection and localization instability \citep{wang2025lsod,nikouei2025small}. Consequently, UAV small-object detection depends on two complementary capabilities: exploiting wide-area context to distinguish targets from background clutter and preserving fine-grained spatial structures for precise localization.

Among deep-learning-based detectors, the YOLO family is widely used in real-time applications because of its compact one-stage architecture and favorable accuracy--efficiency balance \citep{wang2023yolov7,wang2023goldyolo,tian2025yolov12}. UAV-oriented variants have mainly improved local feature extraction and multi-scale fusion. For example, \citet{zhao2024dense} introduced scene-context-guided fusion, while \citet{wang2024adaptive} developed an adaptive receptive-field enhancement network. These CNN-based designs preserve local spatial inductive bias and are effective when target boundaries and neighborhood patterns remain distinguishable. However, when sparse target responses are embedded in complex wide-area backgrounds, strengthening local operators alone may not provide sufficient context for reliable target--background discrimination. Vision Transformers address this limitation through global self-attention \citep{dosovitskiy2021vit,liu2021swin}, but their quadratic complexity \(\mathcal{O}(N^2)\) becomes costly for high-resolution aerial features. The resulting challenge is therefore not simply to improve local feature extraction, but to introduce affordable long-range perception without sacrificing the spatial structures required for tiny-object localization.

To achieve long-range dependency modeling without the quadratic complexity of self-attention, State Space Models (SSMs) have recently attracted increasing attention. \citet{gu2024mamba} proposed Mamba with linear-time sequence modeling capability. Subsequently, \citet{liu2024vmamba} adapted the SSM paradigm to visual representation learning through selective scanning over two-dimensional image structures, while \citet{zhu2024visionmamba} further explored efficient visual representation with bidirectional state-space modeling. To reduce scanning redundancy and improve representation efficiency under high-resolution visual inputs, \citet{pei2025efficientvmamba} proposed EfficientVMamba with atrous-based selective scanning, and \citet{shaker2025groupmamba} introduced GroupMamba with group-based state-space modeling. These developments establish SSMs as a practical alternative for efficient long-range visual modeling.

Building upon these advances, recent studies have extended state-space modeling from visual representation learning to YOLO-style detection and multi-scale feature aggregation. KSCNet introduces SSM/VSS modules into the neck for multi-scale semantic interaction \citep{li2025kscnet}, while Mamba-YOLO employs detection-oriented state-space blocks across both backbone feature extraction and neck aggregation \citep{wang2025mamba}. These studies demonstrate that state-space modeling can be effectively incorporated into different stages of modern detectors and provide an alternative mechanism for long-range contextual interaction.

For UAV small-object detection, however, contextual propagation and local spatial reconstruction constitute different functional requirements. The backbone benefits from broad contextual interaction for distinguishing weak targets from large-area background interference, whereas repeated FPN/PAN aggregation places greater emphasis on explicit local neighborhoods, boundary-sensitive responses, and two-dimensional spatial continuity. How state-space perception and convolutional reconstruction should be allocated according to these stage-specific requirements remains comparatively less explored. Cross-scale fusion introduces a further correspondence problem because fixed interpolation cannot adapt its sampling locations to the current feature content \citep{liu2023dysample}. The resulting multi-scale representations must finally support classification and localization, which exhibit different response preferences \citep{dai2021dynamic,feng2021tood}. These considerations motivate a detector-level design that coordinates contextual perception, local reconstruction, cross-scale alignment, and task interaction as successive but complementary functions.

To address these requirements, this paper proposes Hybrid Dual-domain Mamba-YOLO (HDMamba-YOLO), an efficient UAV small-object detector based on a stage-wise heterogeneous SSM--CNN architecture. An EfficientVMamba-based EVSS backbone first establishes long-range contextual perception over high-resolution UAV features \citep{pei2025efficientvmamba}. At the perception--reconstruction interface, DST-Wrapper re-encodes the deepest backbone representation, after which Native C3k2-ASSAF performs progressive local two-dimensional reconstruction throughout repeated FPN/PAN aggregation. As the reconstructed features are propagated and fused across pyramid levels, DySample performs content-adaptive resampling to improve cross-resolution spatial correspondence \citep{liu2023dysample}. The aligned multi-scale features are subsequently delivered to OS-CVTIA, which coordinates the distinct contextual and boundary-sensitive information required by classification and localization. These successive stages constitute the perception--reconstruction--alignment--interaction pipeline of HDMamba-YOLO.

The main contributions of this work are summarized as follows:
\begin{enumerate}
    \item We formulate a stage-wise heterogeneous SSM--CNN architecture for UAV small-object detection. Rather than extending a homogeneous operator family throughout the detector, HDMamba-YOLO concentrates EfficientVMamba-derived state-space processing in the relatively high-resolution backbone stages for efficient long-range contextual perception, while retaining convolutional FPN/PAN aggregation in the neck for explicit local two-dimensional reconstruction. PhasePatchMerging2D implements phase-aware hierarchical transitions between the state-space stages, thereby aligning the operator allocation with the distinct representational requirements of different detection stages.

    \item We develop an ASSAF-centered reconstruction pathway to preserve boundary-sensitive local structures and incorporate spatially bounded contextual information during repeated multi-scale aggregation. The proposed ASSAF jointly models directionally factorized local details and fixed-grid neighborhood context, and adaptively calibrates the two responses through independent sigmoid gates. Based on this operator, DST-Wrapper performs transition-oriented re-encoding at the backbone--neck interface, whereas Native C3k2-ASSAF embeds progressive reconstruction into the original C2f/C3k2 split--iterate--concatenate topology without discarding its feature-reuse structure.

    \item To complete the detector-level processing pipeline, we integrate DySample at the top-down resampling nodes for content-adaptive cross-scale alignment and develop OS-CVTIA for macro--micro receptive-field interaction and asymmetric task-specific modulation before classification and localization. Together with the perception and reconstruction stages, these components instantiate the proposed perception--reconstruction--alignment--interaction pipeline. Extensive experiments on VisDrone2019 and AI-TOD, including progressive, architecture-level, transition-operator, branch-level, alignment, detection-head, and localization-objective ablations, substantiate the stage-wise design rationale and clarify the contributions and limitations of its individual components.
\end{enumerate}

\section{Related Work}

\subsection{General Object Detectors}

Modern object detectors are commonly divided into two-stage and one-stage paradigms. Two-stage methods, represented by Faster R-CNN \citep{ren2017fasterrcnn} and Cascade R-CNN \citep{cai2018cascade}, refine region proposals before classification and box regression, generally providing strong localization accuracy at the cost of a more involved inference pipeline. One-stage methods, including SSD \citep{liu2016ssd}, RetinaNet \citep{lin2017retinanet}, and the YOLO family \citep{redmon2016yolo}, formulate detection as dense prediction and provide a favorable accuracy--efficiency trade-off for real-time applications. Within the YOLO lineage, architectural progress has been closely associated with more efficient gradient propagation and feature aggregation. YOLOv4 and YOLOv5 incorporate CSP-style connectivity \citep{bochkovskiy2020yolov4,jocher2020yolov5,wang2020cspnet}, YOLOv8 adopts the C2f topology for richer feature reuse \citep{jocher2023yolov8}, and YOLOv10 introduces end-to-end, NMS-free prediction \citep{wang2024yolov10}. YOLOv11 further employs lightweight C3k2-style blocks and deep spatial-semantic processing \citep{khanam2024yolov11}. Related designs such as Gold-YOLO \citep{wang2023goldyolo} and YOLOv12 \citep{tian2025yolov12} likewise emphasize efficient information exchange and contextual modeling.

The locality of convolution nevertheless limits direct interaction between spatially distant regions. Transformer-based detectors address this limitation through global token interaction. DETR reformulates detection as set prediction \citep{carion2020detr}; Deformable DETR restricts attention to sparse samples around reference points \citep{zhu2021deformabledetr}; and RT-DETR decouples intra-scale interaction from cross-scale fusion through a hybrid encoder \citep{zhao2024rtdetr}. Although these designs improve global reasoning, the quadratic complexity of self-attention with respect to the token count remains unfavorable for high-resolution aerial features \citep{dosovitskiy2021vit,liu2021swin}. UAV small-object detection therefore requires an alternative that can enlarge the effective interaction range without discarding the local spatial structures on which precise localization depends.

\subsection{Small-Object Detection in UAV Imagery}

Small-object detection depends strongly on preserving weak responses across feature scales. FPN transfers high-level semantics to shallow feature maps through a top-down pathway \citep{lin2017fpn}, while PANet augments this design with bottom-up aggregation to shorten the information path from low-level localization cues \citep{liu2018panet}. Subsequent UAV-oriented methods have strengthened shallow features, high-frequency responses, and cross-level interaction. Feature-mining strategies recover informative responses that may otherwise be suppressed by background noise \citep{liu2022featuremining}, and HS-FPN explicitly enhances high-frequency and spatial information for tiny-object representation \citep{shi2025hsfpn}. LSOD-YOLO pursues a lightweight balance between representation capacity and deployment efficiency \citep{wang2025lsod}, whereas MFEL-YOLO combines backbone enhancement with efficient path aggregation for UAV aerial imagery \citep{hou2025mfel}. These studies improve multi-scale representation, but repeated fusion can still weaken the local geometric evidence of targets occupying only a few feature locations.

This limitation is especially pronounced in UAV imagery, where small targets are densely distributed, weakly textured, and embedded in heterogeneous backgrounds \citep{wang2021tiny,nikouei2025small}. Reliable detection consequently requires both wide-area context for target--background discrimination and spatially explicit local evidence for localization. Scene-context-guided fusion strengthens contextual support in dense aerial scenes \citep{zhao2024dense}, while adaptive receptive-field enhancement adjusts spatial aggregation to scale variation \citep{wang2024adaptive}. Deformable operators further provide content-dependent sampling for geometrically varying objects \citep{xiong2024dcnv4}. Such flexibility, however, can increase structural and computational overhead, leaving the coordination of contextual modeling, local detail preservation, and lightweight deployment unresolved.

Convolutional studies have approached local structure and contextual aggregation through asymmetric, large-kernel, and selectively calibrated operators. ACNet shows that asymmetric convolutions can strengthen the horizontal and vertical skeletons of convolutional kernels \citep{ding2019acnet}. RepLKNet demonstrates that large depthwise kernels can expand the effective receptive field while retaining regular grid-based processing \citep{ding2022replknet}, and LSKNet adapts large selective kernels to the scale and contextual variation of remote-sensing objects \citep{li2023lsknet}. LSKA further factorizes large-kernel spatial interaction into separable horizontal and vertical operations to reduce computational and memory costs \citep{lau2024lska}. Feature recalibration provides another form of adaptive fusion: SENet predicts channel-wise modulation from globally aggregated statistics \citep{hu2018senet}, whereas SKNet performs competitive selection among receptive-field branches through normalized attention \citep{li2019sknet}. These works establish useful mechanisms for local--contextual representation, but they primarily operate within individual blocks or backbones and do not determine how local reconstruction should be organized throughout repeated FPN/PAN aggregation.

Cross-scale fusion also depends on the resampling operator. Nearest-neighbor and bilinear interpolation use fixed sampling patterns that cannot adapt to spatially varying object structures. CARAFE addresses this issue through content-aware feature reassembly with dynamically generated kernels \citep{wang2019carafe}, whereas DySample formulates upsampling as lightweight point sampling with content-dependent offsets \citep{liu2023dysample}. For UAV targets whose boundaries may span only a few feature locations, content-adaptive resampling offers a practical means of improving correspondence between low-resolution semantic features and high-resolution spatial details. It nevertheless addresses cross-resolution alignment rather than the preceding reconstruction of local geometry.

Classification and localization constitute a further source of representational mismatch. Double-Head R-CNN reports that fully connected and convolutional heads exhibit different preferences for classification and box regression \citep{wu2020doublehead}, while task-aware spatial disentanglement shows that the two tasks may attend to different object regions \citep{song2020tsd}. Dynamic Head unifies scale-, spatial-, and task-aware attention \citep{dai2021dynamic}, and TOOD promotes alignment between classification and localization through task-interactive features and task-aligned learning \citep{feng2021tood}. In UAV scenarios, TPH-YOLOv5 additionally introduces a Transformer prediction head to strengthen contextual reasoning \citep{zhu2021tph}. These methods demonstrate the importance of task-specific adaptation, but a UAV detector must provide such interaction without obscuring weak local evidence or imposing excessive head complexity.

\subsection{State Space Models for Visual Tasks}

State Space Models (SSMs) have emerged as an efficient alternative for long-sequence modeling. S4 introduces a structured state-space formulation for capturing long-range dependencies \citep{gu2022s4}, and Mamba makes the state transition input-dependent while providing hardware-aware selective scanning \citep{gu2024mamba}. Its computational cost grows linearly with sequence length, making state-space modeling attractive when high-resolution visual inputs render dense self-attention expensive.

Visual SSMs adapt sequence scanning to two-dimensional representations. Vision Mamba introduces bidirectional state-space modeling for visual recognition \citep{zhu2024visionmamba}, whereas VMamba employs cross-scan routes to propagate information over two-dimensional feature maps \citep{liu2024vmamba}. EfficientVMamba reduces redundant traversal through atrous selective scanning \citep{pei2025efficientvmamba}, and GroupMamba improves efficiency through group-wise state-space modeling \citep{shaker2025groupmamba}. Other studies explicitly restore spatial inductive bias: Spatial-Mamba incorporates structure-aware state fusion and dilated convolution \citep{xiao2025spatialmamba}, indicating that efficient sequence propagation and explicit spatial processing are complementary rather than interchangeable.

Recent work has further explored heterogeneous visual backbones that combine state-space modeling with other spatial operators. MobileMamba integrates Mamba-based long-range interaction, multi-kernel depthwise convolution, and wavelet-based high-frequency enhancement within a multi-receptive-field feature interaction block \citep{he2025mobilemamba}. MambaVision instead constructs a hierarchical backbone that combines redesigned Mamba mixers with self-attention in later stages \citep{hatamizadeh2025mambavision}. MambaOut examines the necessity of the SSM component from a task-dependent perspective: removing SSM improves its ImageNet classification baseline, whereas the resulting models remain behind the strongest visual Mamba backbones in several COCO detection and ADE20K segmentation comparisons \citep{yu2025mambaout}. Its results therefore suggest that SSM utility is more likely to emerge in long-sequence dense prediction than in conventional short-sequence classification, while stopping short of prescribing where SSMs should be placed within a detector.

State-space modeling has also been incorporated into object detectors. KSCNet coordinates KAN- and VSS-based components for UAV small-object feature extraction and multi-scale aggregation \citep{li2025kscnet}. Mamba-YOLO deploys its detection-oriented ODSSBlock in both the backbone and neck, thereby extending state-space processing across feature extraction and pyramid aggregation \citep{wang2025mamba}. HRMamba-YOLO combines EfficientVMamba-derived processing with a high-resolution feature pyramid for UAV small objects \citep{wu2024hrmamba}, while MambaNeXt-YOLO adopts block-level CNN--Mamba hybridization together with asymmetric multi-branch fusion \citep{lei2025mambanext}. In remote-sensing dense prediction, LGMM-Net likewise coordinates convolutional detail extraction with Mamba-based global modeling \citep{fang2026lgmm}.

Existing visual and detection architectures differ substantially in where state-space operators are introduced: some construct SSM-dominant backbones, some extend SSM processing into pyramid aggregation, and others combine Mamba with convolution or attention within individual blocks. MobileMamba and MambaVision demonstrate operator complementarity at the block or backbone level, whereas MambaOut cautions against treating SSM as universally necessary. What remains insufficiently examined is detector-stage allocation: whether the same state-space operator should be extended through repeated FPN/PAN fusion, or whether contextual perception and local two-dimensional reconstruction should be assigned to different stages according to their functional requirements.

\subsection{Synthesis and Research Gap}
\label{sec:research_gap}

For UAV small-object detection, the preceding findings can be organized according to four stage-specific representational requirements. First, weak target responses require broad contextual perception to distinguish objects from structurally similar background clutter, yet dense self-attention is costly on high-resolution aerial features. Second, multi-scale aggregation must preserve or reconstruct explicit local neighborhoods, boundary-sensitive responses, and two-dimensional continuity after hierarchical feature extraction. Third, the reconstructed representations must be transferred across pyramid levels without introducing avoidable spatial mismatch. Finally, the aligned multi-scale features must be adapted to the different semantic and geometric preferences of classification and localization.

Existing methods provide effective mechanisms for individual requirements, including efficient state-space scanning, local and large-kernel spatial operators, adaptive resampling, and task-aware prediction. However, these mechanisms are commonly studied as isolated module improvements, within-block hybrids, or homogeneous replacements across multiple detector stages. The central gap is therefore not the absence of another feature-enhancement operator, but the lack of a detector-level principle for coordinating complementary operators. In particular, long-range state-space interaction is potentially valuable where high-resolution features still retain weak target evidence, whereas repeated FPN/PAN aggregation requires explicit local two-dimensional processing to reconstruct geometry across semantic levels. Cross-scale propagation then calls for adaptive alignment, and the prediction stage requires lightweight task-specific interaction. This analysis motivates the perception--reconstruction--alignment--interaction formulation developed in the following section, in which each operator is assigned according to the dominant representational requirement of its detector stage.


\section{Proposed Method}
\label{sec:proposed_method}

This section presents the proposed Hybrid Dual-domain Mamba-YOLO (HDMamba-YOLO). Following the functional requirements summarized in Section~\ref{sec:research_gap}, the method is described through four successive stages: efficient state-space perception, ASSAF-centered local spatial reconstruction, content-adaptive cross-scale alignment, and lightweight classification--localization interaction.

\subsection{Overview}
\label{sec:method_overview}

HDMamba-YOLO is organized around four successive functions: perception, reconstruction, alignment, and interaction. These stages impose different representational requirements in UAV small-object detection. HDMamba-YOLO therefore adopts a stage-wise heterogeneous SSM--CNN design: state-space modeling is used to establish long-range contextual perception before substantial spatial compression, CNN-style operators perform explicit local two-dimensional reconstruction during pyramid aggregation, DySample refines cross-scale spatial correspondence, and OS-CVTIA introduces task-specific interaction before prediction.

UAV tiny objects often provide limited local appearance and weak boundary evidence. At relatively high feature resolutions, broader scene context can therefore complement local cues before repeated downsampling reduces spatial detail. HDMamba-YOLO uses EfficientVMamba-based EVSS stages for this contextual perception and introduces PhasePatchMerging2D at the corresponding stage transitions so that the four spatial sampling phases are made explicit before channel projection.

As the feature pyramid is constructed, the representational emphasis shifts from broad contextual interaction to repeated local aggregation. DST-Wrapper bridges this change of role at the backbone--neck interface, and Native C3k2-ASSAF performs local two-dimensional reconstruction while preserving the split--iterative--concatenate topology of C3k2. This stage-wise allocation couples the long-range dependency modeling of EVSS with explicit CNN-based local reconstruction in the neck, matching the operator type to the representational demand of each stage.

Cross-scale fusion introduces a separate spatial correspondence problem. DySample is therefore used at the two top-down upsampling nodes to generate content-dependent sampling coordinates before lateral fusion. The reconstructed and aligned P3, P4, and P5 features are then processed by OS-CVTIA, which introduces lightweight macro--micro interaction for localization and classification before the native YOLO prediction branches. Together, these operations instantiate the perception--reconstruction--alignment--interaction (P--R--A--I) pipeline.

\begin{figure*}[!t]
    \centering
    \includegraphics[
        width=0.98\textwidth,
        height=0.58\textheight,
        keepaspectratio
    ]{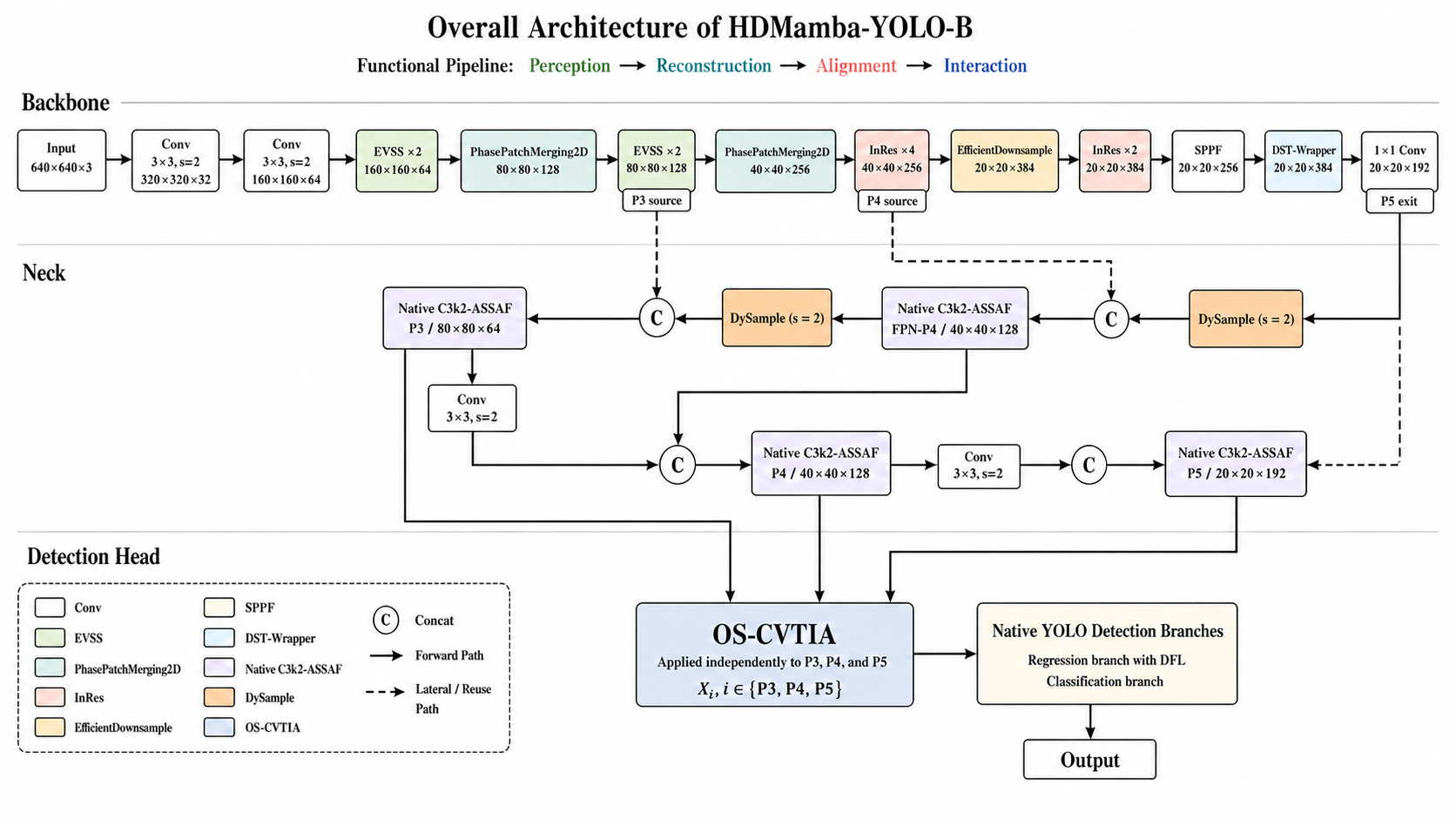}
    \caption{Overall architecture and functional pipeline of HDMamba-YOLO-B. The detector follows a perception--reconstruction--alignment--interaction design: EVSS performs efficient contextual perception and PhasePatchMerging2D constructs phase-aware hierarchical transitions; DST-Wrapper and Native C3k2-ASSAF form an ASSAF-centered local reconstruction pathway; DySample provides content-adaptive cross-scale alignment; and OS-CVTIA performs classification--localization interaction before the native YOLO prediction branches.}
    \label{fig:framework}
\end{figure*}

\begin{figure*}[!t]
    \centering
    \includegraphics[
        width=0.98\textwidth,
        keepaspectratio
    ]{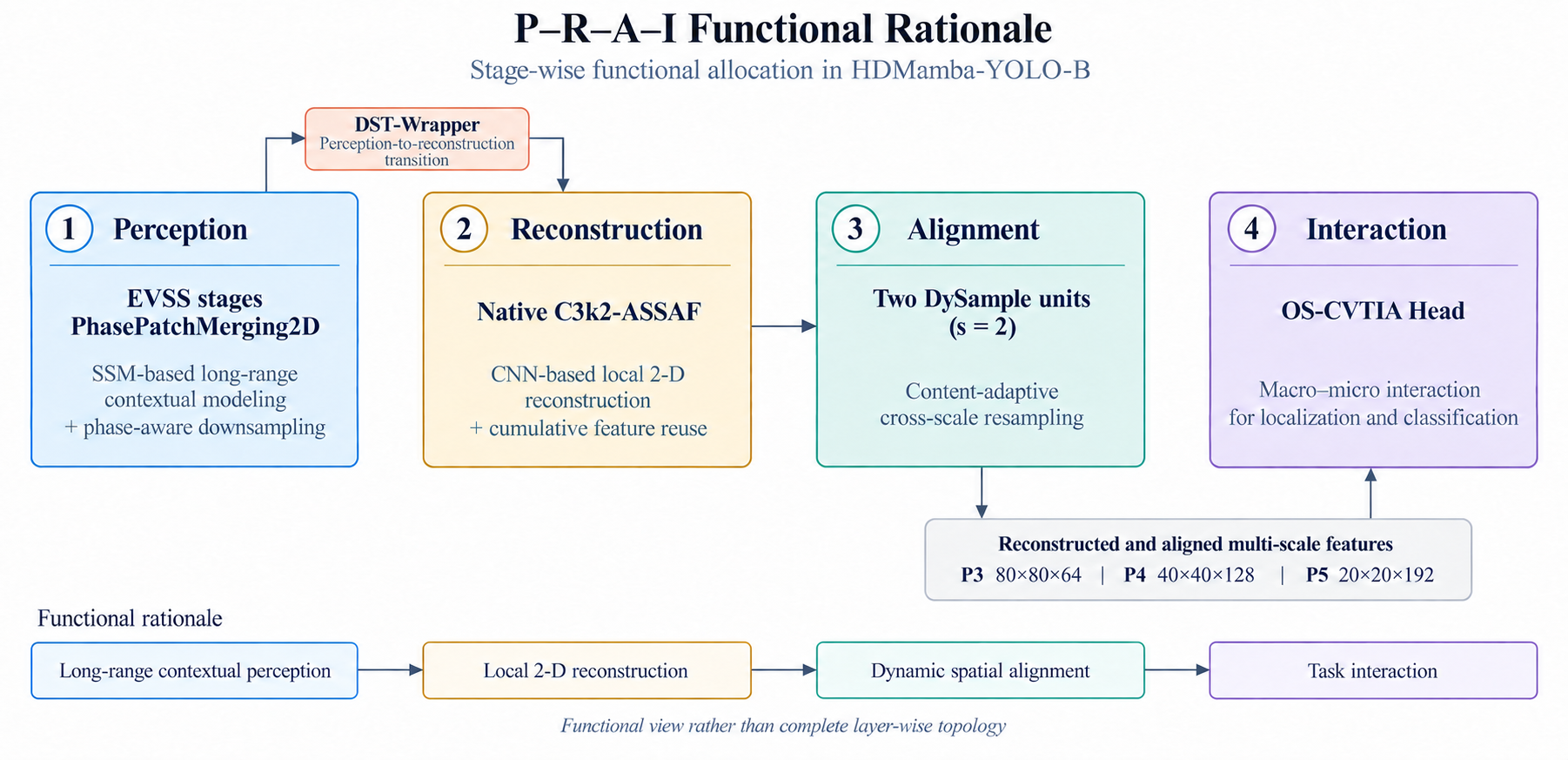}
    \caption{P--R--A--I functional rationale and stage-wise operator allocation in HDMamba-YOLO-B. State-space modeling is assigned primarily to long-range contextual perception, CNN-style processing is retained for local two-dimensional reconstruction, DySample provides content-adaptive cross-scale resampling, and OS-CVTIA performs macro--micro interaction for localization and classification. DST-Wrapper acts as the transition between perception and reconstruction. The reconstructed and aligned multi-scale features are $P3$ ($80\times80\times64$), $P4$ ($40\times40\times128$), and $P5$ ($20\times20\times192$). This figure is a functional abstraction rather than a complete layer-wise topology.}
    \label{fig:prai_rationale}
\end{figure*}

HDMamba-YOLO-B is used as the canonical implementation of the proposed architecture, containing 10.042M parameters and requiring 29.879 GFLOPs at an input resolution of $640\times640$. The compact HDMamba-YOLO-Lite variant contains 5.344M parameters and retains the same stage-wise functional allocation with reduced model capacity; its accuracy--complexity trade-off is examined in the experimental section. Figure~\ref{fig:framework} presents the layer-level architecture, while Fig.~\ref{fig:prai_rationale} isolates the functional rationale underlying the operator allocation.

The architecture in Fig.~\ref{fig:framework} specifies the feature flow, whereas Fig.~\ref{fig:prai_rationale} summarizes why the principal operators are placed at different stages. In this functional view, EVSS and PhasePatchMerging2D provide contextual perception and hierarchical transition, DST-Wrapper connects the backbone to the reconstruction-oriented neck, Native C3k2-ASSAF supplies explicit local two-dimensional reconstruction, DySample performs content-adaptive cross-scale resampling, and OS-CVTIA introduces task interaction at the prediction interface. The diagram is not intended to reproduce every layer connection; it separates the design rationale from the detailed topology shown in Fig.~\ref{fig:framework}.

\subsection{Efficient State-Space Perception}
\label{sec:sequence_domain}

The first functional requirement of HDMamba-YOLO is to establish sufficiently broad contextual perception before tiny-object evidence is progressively compressed by the hierarchical backbone. Conventional convolutional operators provide strong local spatial inductive bias, but their direct feature interactions are inherently concentrated within bounded neighborhoods. This locality can become restrictive in UAV imagery, where tiny targets often contain only weak local appearance and boundary evidence and must be interpreted together with a wider scene context to distinguish them from structurally similar background clutter.

\begin{figure*}[!t]
    \centering
    \includegraphics[
        width=0.72\textwidth,
        keepaspectratio
    ]{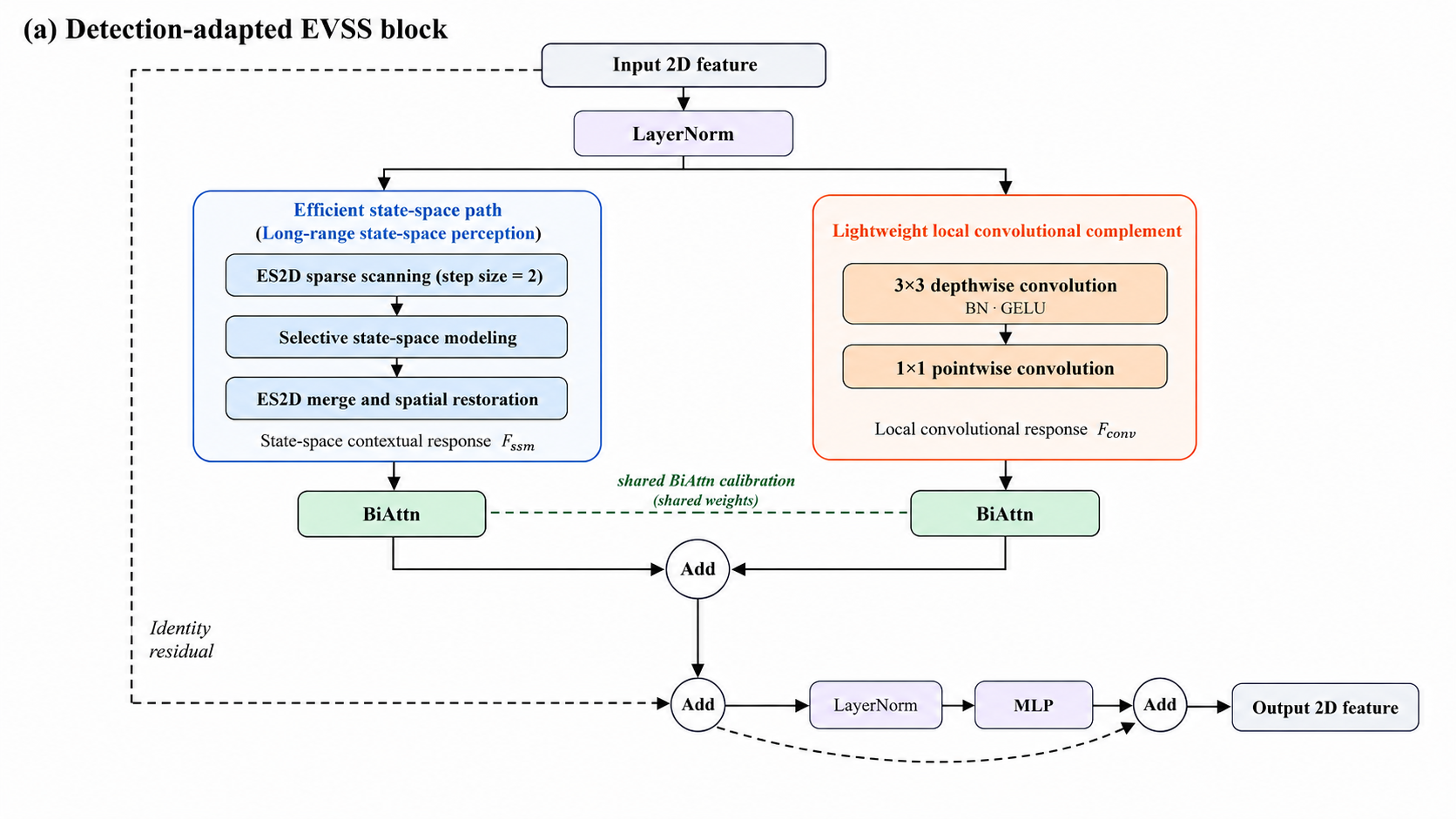}
    \caption{Detection-adapted EVSS block used in the contextual perception stage. The block combines ES2D-based selective state-space propagation with a lightweight local path composed of depthwise convolution, Batch Normalization, GELU, and pointwise projection, followed by shared BiAttn calibration, residual updating, and an MLP transformation.}
    \label{fig:evss_block}
\end{figure*}

HDMamba-YOLO therefore places EfficientVMamba-based EVSS blocks in the relatively high-resolution stages of the YOLO11-S backbone while retaining lightweight convolutional processing in deeper stages. This allocation introduces long-range dependency modeling before substantial spatial compression, where broader contextual information can complement weak local evidence, while the deeper convolutional stages continue efficient local feature refinement. The EVSS structure is shown in Fig.~\ref{fig:evss_block}, and the phase-aware transition used between high-resolution stages is detailed separately in Fig.~\ref{fig:phase_patch_merging}.

\subsubsection{Selective State-Space Modeling with ES2D}
\label{sec:evss_es2d}

EVSS is built upon the selective state-space formulation of Mamba. Given a continuous input signal $x(t)$, a linear state-space system maps it to the output $y(t)$ through a latent state $h(t)$:
\begin{equation}
\frac{\mathrm{d}h(t)}{\mathrm{d}t}
=
\mathbf{A}h(t)
+
\mathbf{B}x(t),
\end{equation}
\begin{equation}
y(t)
=
\mathbf{C}h(t)
+
\mathbf{D}x(t),
\end{equation}
where $\mathbf{A}$ denotes the state-transition matrix, $\mathbf{B}$ and $\mathbf{C}$ represent the input and observation mappings, respectively, and $\mathbf{D}$ denotes the direct feedthrough mapping from the input to the output. After discretization, the state evolution can be expressed as
\begin{equation}
h_t
=
\bar{\mathbf{A}}_t h_{t-1}
+
\bar{\mathbf{B}}_t x_t,
\end{equation}
\begin{equation}
y_t
=
\mathbf{C}_t h_t
+
\mathbf{D}x_t.
\end{equation}

Unlike conventional linear time-invariant SSMs, Mamba makes the discretization and input/output mappings content-dependent, enabling the state transition to selectively preserve or suppress information according to the current input. This selective recurrence provides long-range dependency modeling with computational complexity that grows linearly with the sequence length, making it suitable for high-resolution visual representations.

Applying dense selective scanning directly to high-resolution aerial features would partially offset the efficiency advantage sought in the perception stage. EVSS therefore adopts the Efficient 2D Scanning (ES2D) strategy of EfficientVMamba \citep{pei2025efficientvmamba}, which reorganizes the two-dimensional feature map into complementary sparse scanning groups and applies selective state-space propagation within these groups.

Let $X\in\mathbb{R}^{H\times W\times C}$ denote a two-dimensional feature map. The ES2D transformation can be abstractly expressed as
\begin{equation}
\{X_k\}_{k=1}^{K}
=
\mathcal{S}_{\mathrm{ES2D}}(X),
\end{equation}
where $\mathcal{S}_{\mathrm{ES2D}}(\cdot)$ denotes the sparse spatial decomposition. With the scanning step set to 2, $K=4$, and $X_k$ denotes the $k$-th complementary phase-aware sparse scanning group. Selective state-space modeling is then performed independently on each group:
\begin{equation}
Z_k
=
\operatorname{SSM}(X_k),
\qquad k=1,\ldots,K,
\end{equation}
and the resulting responses are restored to the original spatial organization through
\begin{equation}
Y
=
\mathcal{R}_{\mathrm{ES2D}}
\left(
Z_1,\ldots,Z_K
\right),
\end{equation}
where $\mathcal{R}_{\mathrm{ES2D}}(\cdot)$ denotes spatial restoration. By reducing redundant dense traversal while maintaining complementary spatial coverage, ES2D enables EVSS to establish broad contextual interactions at relatively high feature resolutions with limited computational overhead.

Importantly, EVSS does not rely on the state-space branch alone. After LayerNorm, the normalized input is processed in parallel by the ES2D state-space path and a lightweight local convolutional path. The latter applies a $3\times3$ depthwise convolution followed by Batch Normalization and GELU, and then uses a $1\times1$ pointwise projection for channel mixing. Let $F_{\mathrm{ssm}}$ and $F_{\mathrm{conv}}$ denote the outputs of the state-space and local convolutional paths, respectively. Their interaction can be summarized as
\begin{equation}
F_{\mathrm{mix}}
=
\mathcal{A}(F_{\mathrm{ssm}})
+
\mathcal{A}(F_{\mathrm{conv}}),
\end{equation}
where $\mathcal{A}(\cdot)$ denotes the BiAttn calibration used in the EVSS block. The mixed representation is then incorporated through residual updating, followed by a feed-forward transformation:
\begin{equation}
X'
=
X+F_{\mathrm{mix}},
\qquad
Y
=
X'
+
\operatorname{MLP}
\left(
\operatorname{LN}(X')
\right).
\end{equation}
Thus, EVSS couples efficient long-range state-space propagation with a lightweight local convolutional path, combining broad contextual interaction with local spatial refinement within the same block.

\subsubsection{Phase-Aware Hierarchical Transition}
\label{sec:phase_patch_merging}

Object detection requires hierarchical representations at progressively reduced spatial resolutions. Therefore, the contextual features produced by the high-resolution EVSS stages must be downsampled before being propagated to deeper semantic levels. A conventional stride-2 convolution couples spatial decimation and channel transformation in a single operation, which may discard part of the phase-specific local evidence before cross-channel fusion. This issue is particularly relevant to UAV tiny objects, whose discriminative structures may occupy only a few spatial positions.

PhasePatchMerging2D follows the established Patch Merging principle of rearranging the four interleaved samples in each $2\times2$ neighborhood before channel projection, as introduced for hierarchical visual representations in Swin Transformer \citep{liu2021swin} and subsequently used in detection-oriented state-space architectures such as Mamba-YOLO \citep{wang2025mamba}. In HDMamba-YOLO, this operation is adapted to the first two high-resolution EVSS transitions to expose all four sampling phases to joint normalization and channel projection before dimensionality reduction. Its role is therefore a detection-stage adaptation and placement of an established hierarchical transition, rather than the introduction of a new patch-merging principle.

\begin{figure*}[!t]
    \centering
    \includegraphics[
        width=0.98\textwidth,
        keepaspectratio
    ]{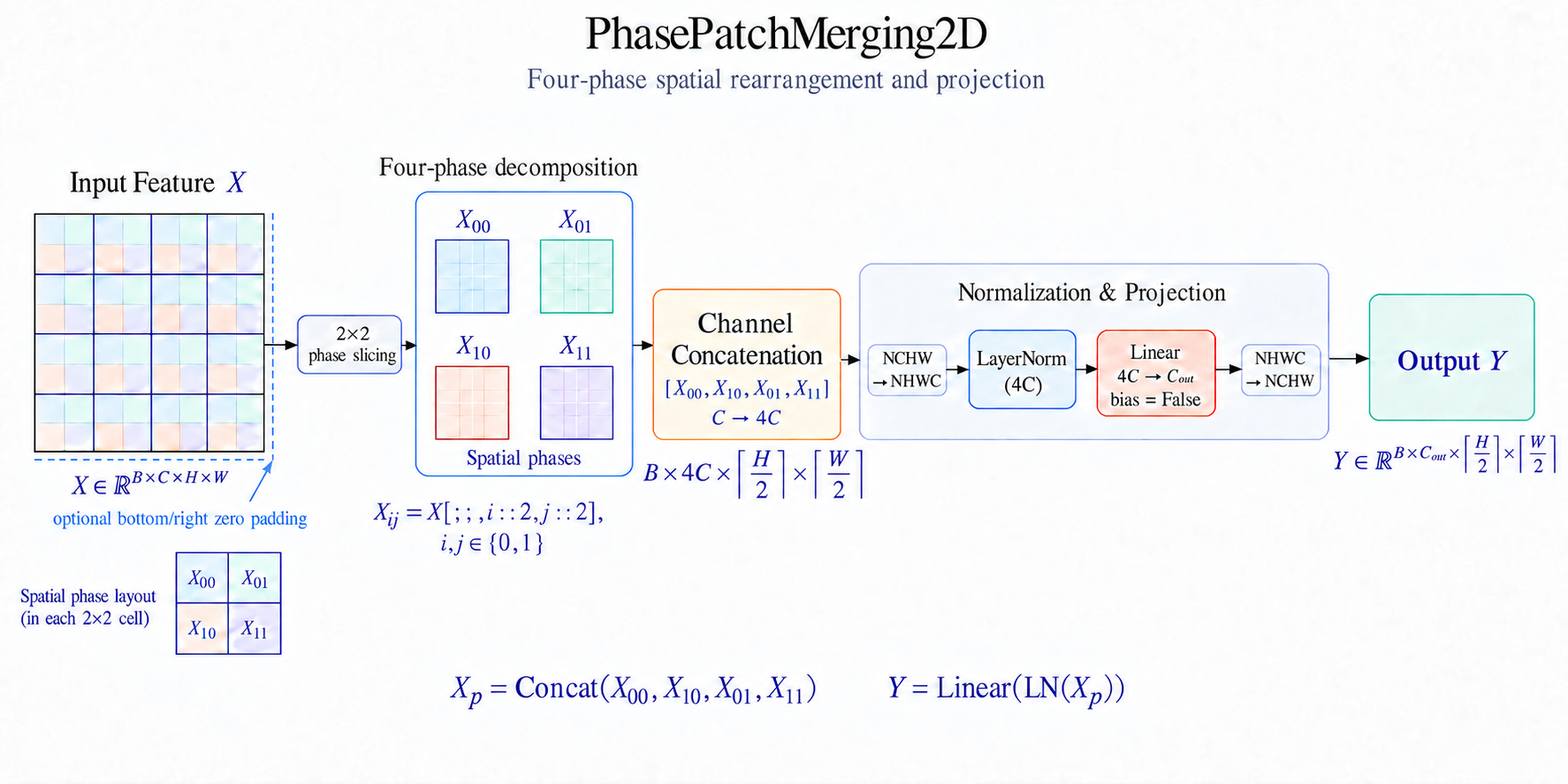}
    \caption{Mechanism of PhasePatchMerging2D. The input feature is decomposed into four parity-defined spatial phases, concatenated in the implementation order $[X_{00},X_{10},X_{01},X_{11}]$, converted to channel-last layout for LayerNorm and bias-free linear projection, and then returned to channel-first layout. Optional zero padding is applied only to the bottom and/or right boundary when an input dimension is odd.}
    \label{fig:phase_patch_merging}
\end{figure*}

The adapted mechanism is illustrated in Fig.~\ref{fig:phase_patch_merging}. PhasePatchMerging2D first performs optional bottom/right zero padding when needed, decomposes the feature into four spatial phases in a fixed implementation order, concatenates them along the channel dimension, and then applies LayerNorm followed by a bias-free linear projection. No activation or residual branch is introduced inside this transition.

Let
\begin{equation}
X\in\mathbb{R}^{B\times C\times H\times W}
\end{equation}
denote the input feature. If $H$ or $W$ is odd, zero padding is applied only to the bottom or right boundary, respectively, producing an even-sized feature $\widetilde{X}$. The feature is then decomposed according to the row and column parity of each spatial location:
\begin{equation}
X_{pq}[:,:,i,j]
=
\widetilde{X}[:,:,2i+p,\,2j+q],
\qquad
p,q\in\{0,1\}.
\end{equation}
Accordingly, four complementary spatial phases,
$X_{00}$, $X_{10}$, $X_{01}$, and $X_{11}$, are obtained and concatenated in the fixed order used by the implementation:
\begin{equation}
X_{\mathrm{phase}}
=
\operatorname{Concat}_{c}
\left(
X_{00},
X_{10},
X_{01},
X_{11}
\right).
\end{equation}
The resulting representation satisfies
\begin{equation}
X_{\mathrm{phase}}
\in
\mathbb{R}^{
B\times4C\times
\lceil H/2\rceil\times
\lceil W/2\rceil
}.
\end{equation}

For each merged spatial position $(i,j)$, the corresponding four-phase vector can be written as
\begin{equation}
z_{i,j}
=
\left[
x_{2i,2j};
x_{2i+1,2j};
x_{2i,2j+1};
x_{2i+1,2j+1}
\right]
\in\mathbb{R}^{4C}.
\end{equation}
PhasePatchMerging2D subsequently performs channel-wise layer normalization followed by a bias-free linear projection:
\begin{equation}
y_{i,j}
=
W_{r}\operatorname{LN}(z_{i,j}),
\qquad
W_{r}
\in
\mathbb{R}^{C_{\mathrm{out}}\times4C}.
\end{equation}
The final output is therefore
\begin{equation}
Y
\in
\mathbb{R}^{
B\times C_{\mathrm{out}}\times
\lceil H/2\rceil\times
\lceil W/2\rceil
}.
\end{equation}

By explicitly collecting the four spatial phases before dimensionality reduction, PhasePatchMerging2D makes all sampling phases within each $2\times2$ neighborhood jointly available to the subsequent normalization and channel projection. Layer normalization and the subsequent linear projection then perform cross-phase fusion and channel adaptation for the next backbone stage. This phase-aware transition is used between the two high-resolution EVSS stages to construct the hierarchical representation without relying on direct stride-based spatial filtering.

Together, EVSS and PhasePatchMerging2D establish the contextual component of the backbone while constructing the feature hierarchy required by the detector. The next stage has a different emphasis: repeated FPN/PAN aggregation combines features across semantic depths and spatial resolutions, making explicit local neighborhoods, boundary-sensitive responses, and two-dimensional spatial continuity increasingly important.

HDMamba-YOLO therefore adopts a heterogeneous stage-wise allocation. State-space modeling is concentrated in backbone-level contextual perception, while CNN-style aggregation is retained in the neck to provide explicit local neighborhoods, boundary-sensitive responses, and two-dimensional spatial reconstruction. This division of labor motivates the ASSAF-centered reconstruction pathway described next and is summarized by the P--R--A--I rationale in Fig.~\ref{fig:prai_rationale}.

\subsection{ASSAF-Centered Local Spatial Reconstruction}
\label{sec:spatial_reconstruction}

Once the backbone has established broad contextual interactions, the neck must repeatedly combine features from different semantic depths and spatial resolutions. At this point, preserving local neighborhoods, boundary-sensitive responses, and two-dimensional continuity becomes particularly important because tiny-object evidence can be diluted during repeated fusion.

\begin{figure*}[!t]
    \centering

    \begin{subfigure}[t]{0.62\textwidth}
        \centering
        \includegraphics[width=\linewidth]{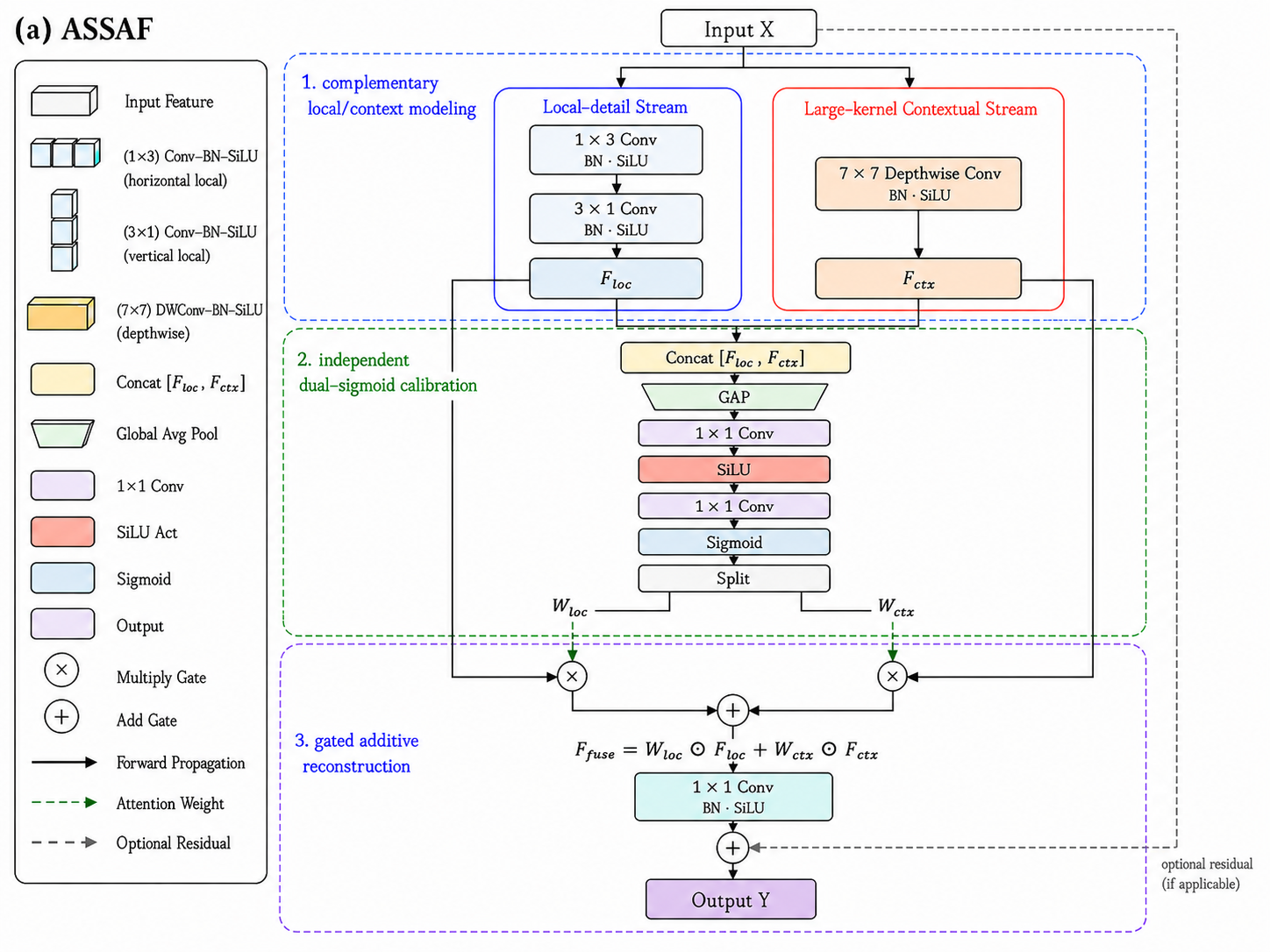}
        \phantomsubcaption
        \label{fig:assaf_module}
    \end{subfigure}

    \vspace{0.7em}

    \begin{subfigure}[t]{0.44\textwidth}
        \centering
        \includegraphics[width=\linewidth]{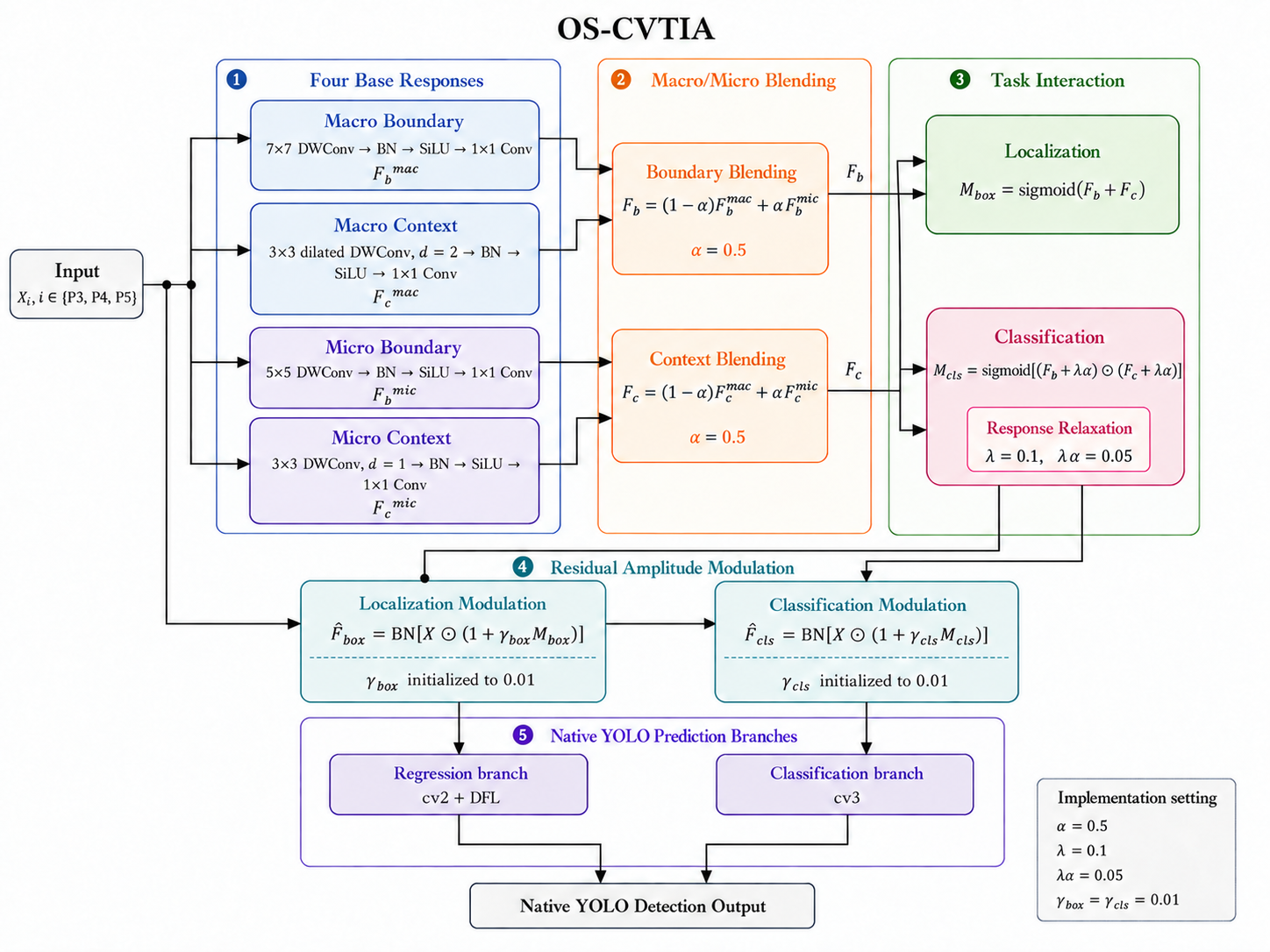}
        \phantomsubcaption
        \label{fig:dst_wrapper}
    \end{subfigure}
    \hfill
    \begin{subfigure}[t]{0.44\textwidth}
        \centering
        \includegraphics[width=\linewidth]{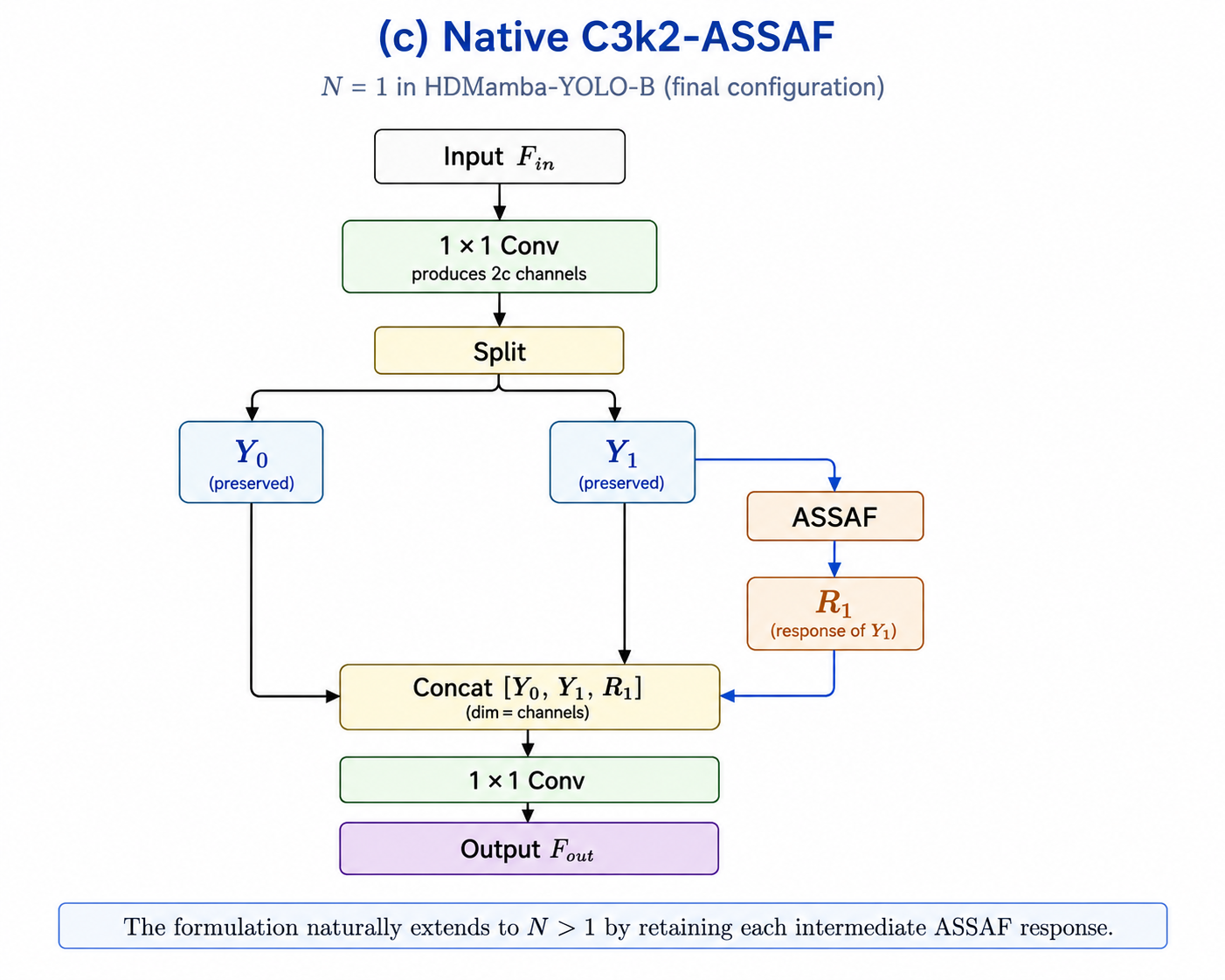}
        \phantomsubcaption
        \label{fig:c3k2_assaf}
    \end{subfigure}

    \caption{ASSAF-centered local spatial reconstruction pathway.
    (a) ASSAF combines directionally factorized local-detail modeling and
    fixed-grid large-kernel contextual modeling with independent dual-sigmoid
    calibration and gated additive reconstruction.
    (b) DST-Wrapper reuses the ASSAF mechanism to re-encode the deepest
    backbone feature at the perception--reconstruction interface.
    (c) Native C3k2-ASSAF embeds ASSAF into the native C2f/C3k2
    split--iterate--concatenate topology and retains the initial partitions
    and intermediate ASSAF responses for progressive feature reuse; the
    canonical HDMamba-YOLO-B configuration uses $N=1$.}
    \label{fig:reconstruction_pathway}
\end{figure*}

HDMamba-YOLO therefore retains a CNN-style aggregation pathway and introduces an ASSAF-centered reconstruction mechanism. DST-Wrapper re-encodes the deepest backbone representation at the backbone--neck interface, after which Native C3k2-ASSAF embeds ASSAF into the original C2f/C3k2 split--iterative--concatenate scaffold for repeated reconstruction and feature reuse along the FPN/PAN pathways. This stage-wise allocation allows EVSS to establish broad contextual dependencies in the backbone, while Native C3k2-ASSAF explicitly reconstructs local two-dimensional structure during repeated FPN/PAN fusion. The reconstruction pathway and its two structural instantiations are illustrated in Fig.~\ref{fig:reconstruction_pathway}.

\subsubsection{Adaptive Spatial-Semantic Attention Fusion}
\label{sec:assaf}

Retaining a CNN-style reconstruction neck preserves local spatial processing, but conventional homogeneous bottlenecks still apply nearly identical transformations to boundary-sensitive details and broader contextual responses. For UAV tiny objects, these two types of information play complementary roles: directional local structures are important for preserving weak contours and spatial continuity, whereas a broader neighborhood response helps distinguish target evidence from structurally similar background patterns. Processing both forms of information through a single transformation may therefore limit the selectivity of local reconstruction.

To address this limitation, Adaptive Spatial-Semantic Attention Fusion (ASSAF) explicitly models the two forms of spatial evidence within a lightweight reconstruction operator. It combines (i) directionally factorized local modeling for short-range spatial continuity, (ii) fixed-grid bounded contextual aggregation for a coordinate-consistent neighborhood reference, (iii) independent dual-sigmoid calibration that allows the two responses to vary without a sum-to-one constraint, and (iv) a conditional residual route when input and output dimensions match. Given $X\in\mathbb{R}^{B\times C\times H\times W}$, these components provide complementary local and contextual responses for subsequent reconstruction.

The local-detail stream is designed to model short-range spatial continuity and direction-sensitive structures that are important for tiny-object localization. Instead of applying a dense $3\times3$ transformation directly, ASSAF factorizes the spatial interaction into sequential horizontal and vertical operations using $1\times3$ and $3\times1$ convolutions. Asymmetric convolutional blocks have been shown to strengthen horizontal and vertical kernel skeletons while retaining two-dimensional spatial support \citep{ding2019acnet}. Motivated by this property, their sequential composition in ASSAF provides an effective $3\times3$ support while explicitly decomposing horizontal and vertical interactions. The resulting factorization offers a compact operator for modeling anisotropic contour transitions and local spatial continuity at repeated reconstruction nodes.

\begin{equation}
\mathbf{F}_{loc}
=
\phi
\left(
\mathcal{B}
\left(
\mathrm{Conv}_{3 \times 1}
\left(
\phi
\left(
\mathcal{B}
\left(
\mathrm{Conv}_{1 \times 3}(X)
\right)
\right)
\right)
\right)
\right),
\end{equation}
where $\mathcal{B}$ denotes Batch Normalization and $\phi$ represents the SiLU activation function.

For receptive-field analysis, ignoring the intervening normalization and nonlinearity, the two directional spatial operators can be regarded as a factorized composition:
\begin{equation}
K_{\mathrm{eff}}
=
K_{3\times1} * K_{1\times3},
\end{equation}
whose spatial support covers a $3\times3$ neighborhood while preserving explicit horizontal--vertical factorization.

The contextual stream is designed to provide a broader yet spatially bounded reference without introducing input-dependent sampling coordinates. Large-kernel convolutions provide a direct mechanism for expanding the effective receptive field in CNNs \citep{ding2022replknet}, and adaptive large-kernel designs have further demonstrated the value of multi-scale contextual modeling in remote-sensing object detection \citep{li2023lsknet}. Drawing on this general motivation, ASSAF employs a fixed-grid $7\times7$ depthwise convolution as a lightweight bounded-context branch:
\begin{equation}
\mathbf{F}_{ctx}
=
\phi
\left(
\mathcal{B}
\left(
\mathrm{DWConv}_{7\times7}(X)
\right)
\right).
\end{equation}
For channel $c$ and spatial position $p$, the pre-normalization depthwise response can be expressed as
\begin{equation}
u_{ctx}^{c}(p)
=
\sum_{q\in\Omega_{7}}
k_c(q)\,x_c(p+q),
\end{equation}
where
$\Omega_{7}=\{-3,\ldots,3\}\times\{-3,\ldots,3\}$
denotes the fixed spatial support of the $7\times7$ kernel.

Unlike offset-dependent dynamic sampling, the sampling coordinates $p+q$ are independent of the input feature. Consequently, feature perturbations do not introduce an additional offset-induced coordinate perturbation into this branch, providing a coordinate-consistent neighborhood reference for local reconstruction. Meanwhile, the contextual dependency remains spatially bounded:
\begin{equation}
\frac{\partial u_{ctx}^{c}(p)}
{\partial x_c(r)}
=
0,
\qquad
r\notin p+\Omega_{7}.
\end{equation}
Thus, each location incorporates a broader neighborhood without allowing arbitrarily distant background responses to directly participate in a single contextual transformation. Because the operation is depthwise, spatial context is extracted independently within each channel before subsequent adaptive calibration and channel projection. This decouples spatial context extraction from cross-channel interaction while maintaining limited computational overhead.

The two streams encode complementary spatial properties and should not be forced to contribute with a fixed proportion at every channel. ASSAF therefore performs content-dependent channel calibration before additive fusion, allowing the contributions of local-detail and contextual responses to vary independently according to the current feature representation. Following the channel-recalibration principle established by squeeze-and-excitation modeling \citep{hu2018senet}, the local and contextual features are first concatenated along the channel dimension and summarized by global average pooling:
\begin{equation}
\mathbf{z}
=
\mathrm{GAP}
\left(
\mathrm{Concat}
\left[
\mathbf{F}_{loc}, \mathbf{F}_{ctx}
\right]
\right).
\end{equation}
A lightweight gating function $\mathcal{G}(\cdot)$ first produces $2C$ channel logits. An element-wise Sigmoid is then applied without cross-branch normalization, and the resulting responses are split into two $C$-dimensional modulation vectors:
\begin{equation}
\left[
\mathbf{W}_{loc},
\mathbf{W}_{ctx}
\right]
=
\mathrm{Split}
\left(
\sigma
\left(
\mathcal{G}
\left(
\mathbf{z}
\right)
\right)
\right),
\end{equation}
where $\sigma(\cdot)$ denotes the Sigmoid function. Because the activation is applied element-wise to the $2C$ outputs of $\mathcal{G}(\cdot)$ before splitting, the resulting $\mathbf{W}_{loc}$ and $\mathbf{W}_{ctx}$ constitute two branch-specific gates rather than a competitively normalized pair. Their channel-wise entries satisfy
\begin{equation}
W_{loc}^{c}\in(0,1),
\qquad
W_{ctx}^{c}\in(0,1),
\qquad
c=1,\ldots,C.
\end{equation}
No cross-branch normalization is imposed between the two modulation vectors. This design differs from the competitive softmax selection used to distribute relative attention across kernel branches in Selective Kernel Networks \citep{li2019sknet}. Instead, the two sigmoid gates in ASSAF can strengthen or suppress the local-detail and contextual responses independently, allowing both forms of evidence to remain active when supported by the current feature representation. The final gating projection is zero-initialized, so both modulation branches start from a neutral coefficient of $0.5$ and progressively learn their respective contributions during optimization.

The fused feature is obtained through channel-aligned gated addition:
\begin{equation}
\mathbf{F}_{fuse}
=
\mathbf{W}_{loc}
\odot
\mathbf{F}_{loc}
+
\mathbf{W}_{ctx}
\odot
\mathbf{F}_{ctx},
\end{equation}
where $\odot$ denotes element-wise multiplication. Because the two branches are defined on the same spatial lattice, gated addition performs channel-wise response calibration without introducing an additional spatial transformation. The fused feature is then projected back to the output space:
\begin{equation}
\mathbf{F}_{proj}
=
\phi
\left(
\mathcal{B}
\left(
\mathrm{Conv}_{1 \times 1}
\left(
\mathbf{F}_{fuse}
\right)
\right)
\right).
\end{equation}

When the input and output dimensions are consistent, a residual connection is adopted:
\begin{equation}
Y
=
X
+
\mathbf{F}_{proj}.
\end{equation}
Otherwise,
\begin{equation}
Y
=
\mathbf{F}_{proj}.
\end{equation}

When the input and output dimensions are identical, the residual path retains the original feature as a direct information route while the fused ASSAF response acts as an adaptive reconstruction correction. This prevents the reconstruction transform from becoming the only path through which the original representation must propagate.

Overall, ASSAF integrates four complementary mechanisms: directionally factorized local modeling for short-range spatial structure, fixed-grid bounded contextual modeling for coordinate-consistent neighborhood reference, independent dual-sigmoid calibration for adaptive local--context response regulation, and residual reconstruction for preserving the original representation when dimensionality permits. Together, these mechanisms form the core local reconstruction operator of HDMamba-YOLO.

\subsubsection{DST-Wrapper for Deep Feature Transition and Re-encoding}
\label{sec:dst_wrapper}

In the original YOLO11-S backbone, C2PSA is placed at the terminal stage to refine the deepest representation through attention-based feature interaction. Such a design is reasonable in the original convolution-dominated backbone, where the terminal module further enriches the high-level representation before multi-scale aggregation. In HDMamba-YOLO, however, broad contextual dependencies have already been explicitly modeled by the preceding EVSS stages. Consequently, the functional requirement of the terminal backbone stage changes: rather than continuing the same emphasis on contextual refinement, the deepest representation must be adapted to the subsequent CNN-based reconstruction pathway.

Based on this requirement, we replace the original C2PSA with DST-Wrapper, an ASSAF-derived transition block that re-encodes the deepest representation at the perception--reconstruction interface. DST-Wrapper reuses the dual-stream reconstruction mechanism defined in Section~\ref{sec:assaf}, and its operation can be compactly expressed as
\begin{equation}
X_{dst}
=
\operatorname{ASSAF}_{c_{in}\rightarrow c_{out}}
\left(
X_d
\right),
\end{equation}
where $X_d$ denotes the deepest backbone feature and the subscript indicates the channel adaptation performed by the ASSAF output projection.

In the canonical HDMamba-YOLO-B configuration, \(c_{\mathrm{in}} \neq c_{\mathrm{out}}\) at this interface; therefore, the conditional identity shortcut inside ASSAF is inactive. DST-Wrapper consequently operates as a gated re-encoding transformation rather than a residual reconstruction block. The subsequent $1\times1$ projection further adapts the re-encoded feature to the width required by the FPN/PAN pathway. The effectiveness of replacing the original terminal C2PSA with this transition-oriented design is further evaluated in the progressive ablation study.

\subsubsection{Native C3k2-ASSAF Aggregation in FPN and PAN}
\label{sec:c3k2_assaf}

DST-Wrapper re-encodes the deepest representation only once at the backbone--neck interface, whereas local spatial reconstruction must be maintained throughout the subsequent top-down and bottom-up aggregation pathways. At each fusion node, features from different semantic depths and spatial resolutions are repeatedly combined, making both progressive reconstruction and feature reuse important for preserving weak tiny-object structures during pyramid construction.

To this end, we embed ASSAF into the native C2f/C3k2 scaffold to construct Native C3k2-ASSAF. Instead of reducing the aggregation block to two terminal feature branches, the proposed structure preserves the native split--iterative--concatenate topology and replaces only the internal bottleneck transformations with ASSAF-based reconstruction. This organization retains the two initial feature partitions together with every intermediate ASSAF response, allowing representations at different reconstruction depths to participate directly in the final aggregation.

Given an input feature map $F_{in}$, a single $1\times1$ projection first generates two hidden partitions:
\begin{equation}
[Y_0,Y_1]
=
\operatorname{Split}
\left(
\operatorname{Conv}_{1\times1}(F_{in})
\right).
\end{equation}
The second partition is then progressively reconstructed:
\begin{equation}
R_1
=
\operatorname{ASSAF}(Y_1),
\end{equation}
\begin{equation}
R_i
=
\operatorname{ASSAF}(R_{i-1}),
\qquad
i=2,\ldots,N.
\end{equation}
Rather than retaining only the terminal reconstruction response, all initial and intermediate representations participate in the final fusion:
\begin{equation}
F_{out}
=
\operatorname{Conv}_{1\times1}
\left(
\operatorname{Concat}
[
Y_0,Y_1,R_1,\ldots,R_N
]
\right).
\end{equation}

The accumulated topology exposes multiple reconstruction depths directly to the output projection. This preserves the native feature-reuse pattern of C2f/C3k2 and provides short information and gradient routes from the final fusion to earlier intermediate representations. These properties are particularly relevant to UAV tiny objects, whose weak local structures can otherwise be progressively diluted during repeated top-down and bottom-up FPN/PAN aggregation.

Native C3k2-ASSAF consequently extends ASSAF from a single reconstruction operator into a repeated multi-scale reconstruction scaffold while retaining the native feature-reuse topology of C2f/C3k2. Local reconstruction is therefore maintained throughout both the FPN and PAN pathways rather than being restricted to the backbone--neck interface.

\subsection{Content-Adaptive Cross-Scale Alignment}
\label{sec:dysample}

Local reconstruction does not eliminate the spatial correspondence requirement introduced during cross-scale feature propagation. In the top-down FPN pathway, low-resolution semantic features must be upsampled before fusion with higher-resolution spatial features. Conventional nearest-neighbor interpolation is inexpensive but static and content-agnostic: its fixed sampling rule cannot adapt to spatially varying tiny-object structures. Because tiny targets occupy only a few pixels, even small spatial quantization errors may perturb their boundaries and cross-scale correspondence. HDMamba-YOLO therefore integrates DySample \citep{liu2023dysample} at the two top-down upsampling nodes to generate content-dependent sampling coordinates and refine sub-pixel spatial alignment before lateral fusion.

\begin{figure*}[!t]
    \centering
    \includegraphics[
        width=0.98\textwidth,
        keepaspectratio
    ]{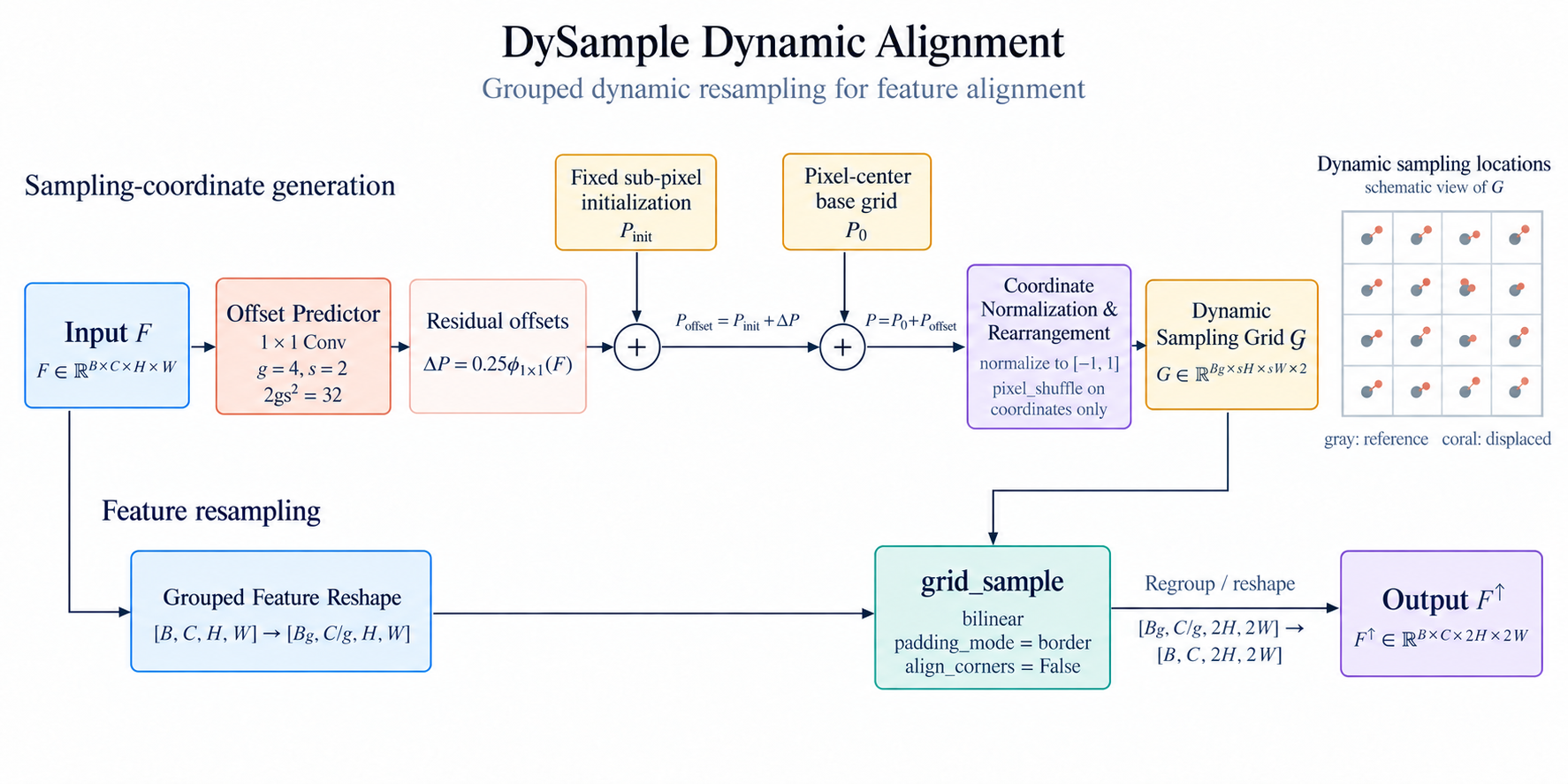}
    \caption{Mechanism of DySample-based dynamic spatial alignment. A $1\times1$ offset predictor generates content-dependent residual offsets, which are combined with fixed sub-pixel initialization and the pixel-center base grid. After coordinate normalization and rearrangement, the resulting dynamic grid resamples grouped feature values using bilinear \texttt{grid\_sample}. In HDMamba-YOLO-B, $g=4$ and $s=2$; \texttt{padding\_mode=border} and \texttt{align\_corners=False} are used.}
    \label{fig:dysample_mechanism}
\end{figure*}

The implementation follows the grouped LP-style formulation of DySample. Figure~\ref{fig:dysample_mechanism} separates its two operations: sampling-coordinate generation and grouped feature resampling. Importantly, the pixel-shuffle operation is applied only to the coordinate representation; the feature values themselves are produced by bilinear \texttt{grid\_sample}.

Let $X\in\mathbb{R}^{B\times C\times H\times W}$ denote the input feature, $g$ the number of channel groups, and $s$ the upsampling factor. The point-wise offset predictor produces $2gs^2$ channels. In the final configuration, $g=4$ and $s=2$, giving 32 offset channels. The residual offsets are
\begin{equation}
\Delta P
=
0.25\,\phi_{1\times1}(X),
\qquad
\Delta P\in
\mathbb{R}^{B\times 2gs^2\times H\times W},
\end{equation}
where $\phi_{1\times1}(\cdot)$ denotes the learned $1\times1$ offset projection. DySample defines a fixed sub-pixel initialization $P_{init}$ and combines it with the learned residual offsets:
\begin{equation}
P_{offset}
=
P_{init}+\Delta P.
\end{equation}
The offset positions are then added to the pixel-center base grid $P_0$:
\begin{equation}
P
=
P_0+P_{offset}.
\end{equation}
Before sampling, the coordinates are normalized to the range required by \texttt{grid\_sample} and rearranged to the target spatial resolution. The resulting grouped sampling grid is
\begin{equation}
G
\in
\mathbb{R}^{Bg\times sH\times sW\times 2},
\end{equation}
where the final dimension stores the two-dimensional sampling coordinates.

In parallel, the input feature is reshaped by groups:
\begin{equation}
X_g
=
\operatorname{Reshape}(X),
\qquad
X_g\in
\mathbb{R}^{Bg\times (C/g)\times H\times W}.
\end{equation}
The upsampled grouped feature is obtained as
\begin{equation}
Y_g
=
\operatorname{GridSample}
\left(
X_g,G
\right),
\end{equation}
using bilinear interpolation with \texttt{padding\_mode=border} and \texttt{align\_corners=False}. Finally, the groups are restored to the original channel organization:
\begin{equation}
Y
=
\operatorname{Regroup}(Y_g),
\qquad
Y\in
\mathbb{R}^{B\times C\times sH\times sW}.
\end{equation}
For the implemented $s=2$ configuration, the spatial resolution is doubled in each dimension.

DySample acts as an alignment operator within the reconstruction neck rather than as an independent representation stage. After top-down resampling and subsequent FPN/PAN aggregation, the reconstructed and aligned P3, P4, and P5 features are passed to the prediction head. Their shared representation is then adapted to the distinct response preferences of localization and classification by OS-CVTIA.

\subsection{Lightweight Classification--Localization Interaction with OS-CVTIA}
\label{sec:os_cvtia}

After reconstruction and cross-scale alignment, the resulting P3, P4, and P5 features are spatially prepared for prediction, but they are still shared by two tasks with different response preferences. Prior studies have shown that classification and localization can favor non-identical feature regions and benefit from task-aware feature or spatial decoupling \citep{wu2020doublehead,song2020tsd,dai2021dynamic,feng2021tood}. In this setting, classification typically benefits from discriminative semantic and contextual evidence, whereas localization places greater emphasis on geometry- and boundary-sensitive responses. Directly passing an identical representation to both prediction branches therefore leaves their distinct feature preferences insufficiently modeled.

To introduce task-specific adaptation without replacing the lightweight YOLO prediction interface, we propose the Omni-Scale Cross-View Task-Interactive Alignment module (OS-CVTIA). For each pyramid feature $X_i$, $i\in\{P3,P4,P5\}$, OS-CVTIA constructs macro- and micro-receptive-field boundary/context responses and blends them using a coefficient $\alpha\in[0,1]$. The resulting representations are subsequently transformed into asymmetric localization and classification masks for task-specific residual modulation.

\begin{figure*}[!t]
    \centering
    \includegraphics[
        width=0.96\textwidth,
        keepaspectratio
    ]{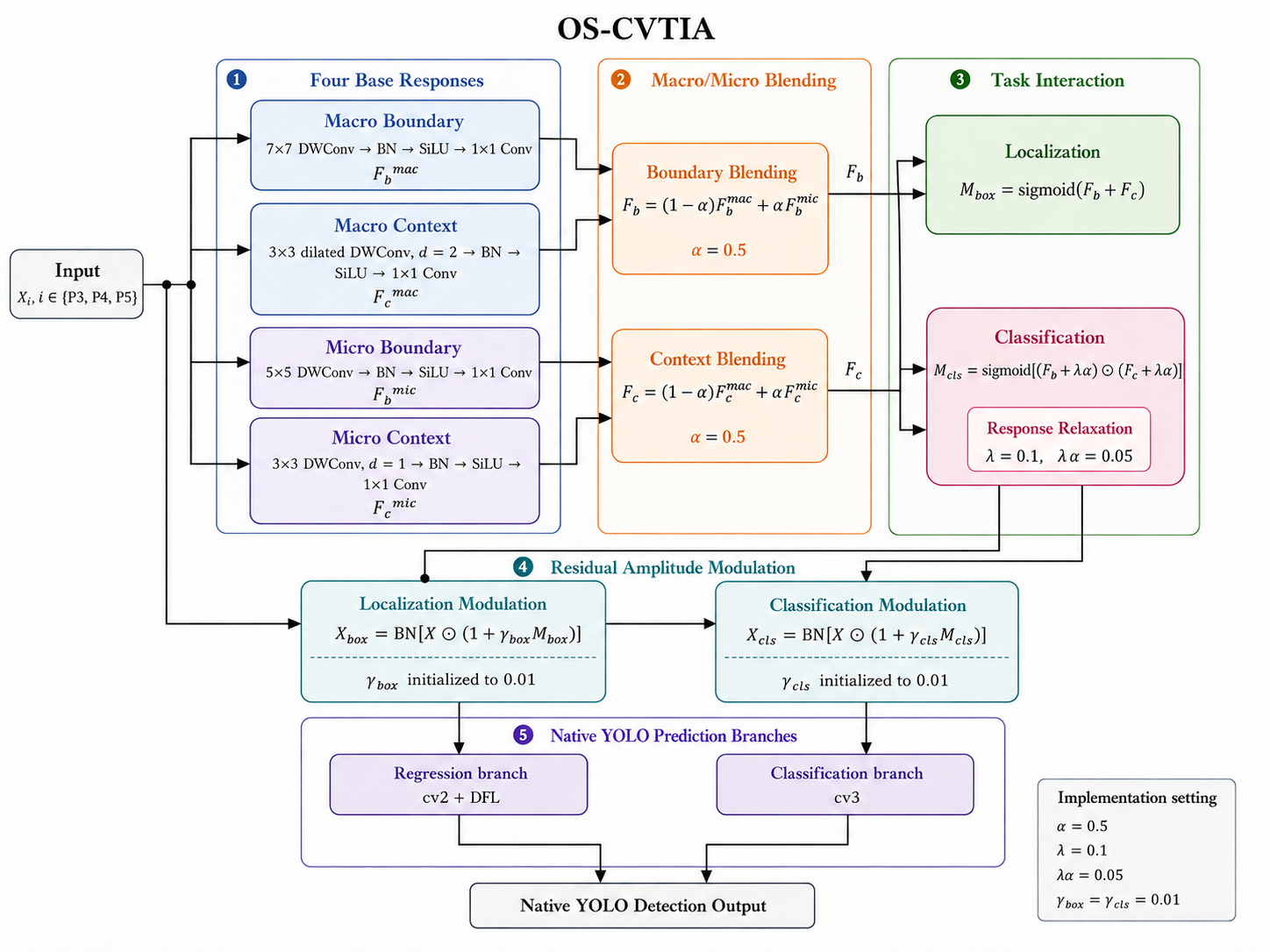}
    \caption{Structure of OS-CVTIA and its native YOLO prediction interface. Four macro/micro boundary--context responses are blended using $\alpha$ and converted into asymmetric localization and classification masks. The masks modulate the shared feature through learnable residual amplitudes before the native regression (\texttt{cv2}+DFL) and classification (\texttt{cv3}) branches. In the final configuration, $\alpha=0.5$, $\lambda=0.1$, $\lambda\alpha=0.05$, and $\gamma_{box}=\gamma_{cls}=0.01$ at initialization.}
    \label{fig:os_cvtia}
\end{figure*}

Given an input feature map $X \in \mathbb{R}^{B \times C \times H \times W}$, the macro stream is designed to provide broader contextual perception. It consists of a boundary-oriented branch with a $7 \times 7$ depthwise convolution and a context-oriented branch with a $3 \times 3$ dilated depthwise convolution. The two macro features are formulated as
\begin{equation}
F_{b}^{mac}
=
\mathcal{P}_{1\times1}
\left(
\phi
\left(
\mathcal{B}
\left(
\mathrm{DWConv}_{7\times7}(X)
\right)
\right)
\right),
\end{equation}
\begin{equation}
F_{c}^{mac}
=
\mathcal{P}_{1\times1}
\left(
\phi
\left(
\mathcal{B}
\left(
\mathrm{DWConv}_{3\times3}^{d=2}(X)
\right)
\right)
\right),
\end{equation}
where $\mathcal{B}$ denotes Batch Normalization, $\phi$ denotes the SiLU activation function, and $\mathcal{P}_{1\times1}$ denotes a $1 \times 1$ pointwise projection. The $7 \times 7$ depthwise convolution provides a broader boundary-sensitive view, while the dilated $3 \times 3$ convolution captures sparse local contextual cues with an enlarged effective receptive field.

In contrast, the micro stream is designed to preserve compact geometric details and weak fine-scale responses. It therefore adopts a more compact receptive-field configuration, replacing the $7\times7$ boundary branch with a $5\times5$ depthwise convolution and using a standard $3\times3$ depthwise convolution without dilation for local context modeling:
\begin{equation}
F_{b}^{mic}
=
\mathcal{P}_{1\times1}
\left(
\phi
\left(
\mathcal{B}
\left(
\mathrm{DWConv}_{5\times5}(X)
\right)
\right)
\right),
\end{equation}
\begin{equation}
F_{c}^{mic}
=
\mathcal{P}_{1\times1}
\left(
\phi
\left(
\mathcal{B}
\left(
\mathrm{DWConv}_{3\times3}^{d=1}(X)
\right)
\right)
\right).
\end{equation}
Together, these operators provide a compact local view that complements the broader boundary and contextual responses of the macro stream before scale modulation.

The macro and micro features are then fused by the continuous scale modulation coefficient $\alpha$:
\begin{equation}
F_{b}
=
(1-\alpha)F_{b}^{mac}
+
\alpha F_{b}^{mic},
\end{equation}
\begin{equation}
F_{c}
=
(1-\alpha)F_{c}^{mac}
+
\alpha F_{c}^{mic}.
\end{equation}
The coefficient $\alpha$ provides continuous control over the macro--micro receptive-field preference. As $\alpha$ approaches $0$, the blended representation is biased toward broader contextual responses; as $\alpha$ approaches $1$, the contribution of compact fine-scale responses increases. This continuous interpolation avoids a hard architectural switch between the two receptive-field ranges. In the canonical HDMamba-YOLO-B implementation, $\alpha$ is fixed to $0.5$, assigning equal nominal contributions to the macro and micro streams before task-specific interaction.

After scale modulation, OS-CVTIA performs asymmetric task interaction for localization and classification. For the localization branch, an additive interaction is adopted to generate the regression modulation mask:
\begin{equation}
M_{box}
=
\sigma
\left(
F_{b}+F_{c}
\right),
\end{equation}
where $\sigma$ denotes the Sigmoid activation function. The additive formulation provides a relaxed interaction in which boundary-oriented and contextual responses can contribute jointly to localization without requiring both responses to be simultaneously strong.

For the classification branch, a stricter multiplicative interaction is adopted. However, directly multiplying two weak responses can over-suppress tiny-object activations. To alleviate this issue, OS-CVTIA introduces an $\alpha$-aware soft-relaxation term into the classification gate:
\begin{equation}
M_{cls}
=
\sigma
\left(
\left(
F_{b}
+
\lambda\alpha
\right)
\odot
\left(
F_{c}
+
\lambda\alpha
\right)
\right),
\end{equation}
where $\odot$ denotes element-wise multiplication and $\lambda$ is fixed to $0.1$. The term $\lambda\alpha$ acts as an $\alpha$-dependent response relaxation rather than as an explicit weighting coefficient for the micro stream. It prevents the multiplicative interaction from becoming excessively restrictive when both responses are weak while retaining the joint boundary--context selectivity of the classification mask. With the final setting $\alpha=0.5$, the relaxation added to each interaction term is $\lambda\alpha=0.05$, with $\lambda=0.1$. Increasing $\alpha$ therefore shifts the blended representation toward the micro stream while simultaneously strengthening the relaxation term. This coupled adjustment is intended to preserve sensitivity to weak fine-scale responses and reduce their excessive suppression by multiplicative classification gating.

Instead of replacing the original feature representation, OS-CVTIA applies residual amplitude modulation to the input feature map. The aligned features for regression and classification are computed as
\begin{equation}
\hat{F}_{box}
=
\mathrm{BN}
\left(
X
\odot
\left(
1+\gamma_{box}M_{box}
\right)
\right),
\end{equation}
\begin{equation}
\hat{F}_{cls}
=
\mathrm{BN}
\left(
X
\odot
\left(
1+\gamma_{cls}M_{cls}
\right)
\right),
\end{equation}
where $\gamma_{box}$ and $\gamma_{cls}$ are learnable scalar modulation coefficients initialized to $0.01$. This small initialization keeps the initial task-specific modulation close to the shared input representation while allowing its amplitude to be progressively adapted during training. The residual formulation preserves the original prediction feature as the main information path, and the subsequent Batch Normalization stabilizes the task-specific feature distributions before prediction.

Finally, the aligned features are fed into the original YOLO detection branches:
\begin{equation}
P_{box}
=
\mathrm{Head}_{box}
\left(
\hat{F}_{box}
\right),
\quad
P_{cls}
=
\mathrm{Head}_{cls}
\left(
\hat{F}_{cls}
\right).
\end{equation}
The modulated features are subsequently processed by the native YOLO regression and classification branches. The regression branch retains the distribution-based box representation and DFL decoding procedure, while the classification branch preserves the native lightweight prediction pathway. OS-CVTIA therefore performs task-specific feature interaction without replacing the established detection formulation.

Through receptive-field blending, asymmetric task masks, and residual amplitude modulation, OS-CVTIA completes the final interaction stage of HDMamba-YOLO. The resulting prediction pipeline therefore follows the functional sequence established throughout this section: Perception $\rightarrow$ Reconstruction $\rightarrow$ Alignment $\rightarrow$ Interaction.

\subsection{Detection Objective}
\label{sec:detection_objective}

HDMamba-YOLO retains the classification and distribution-based regression objectives of the YOLO detection framework. The overall training objective is formulated as
\begin{equation}
\mathcal{L}
=
\lambda_{\mathrm{cls}}\mathcal{L}_{\mathrm{cls}}
+
\lambda_{\mathrm{box}}\mathcal{L}_{\mathrm{CIoU}}
+
\lambda_{\mathrm{dfl}}\mathcal{L}_{\mathrm{DFL}},
\end{equation}
where $\lambda_{\mathrm{cls}}$, $\lambda_{\mathrm{box}}$, and $\lambda_{\mathrm{dfl}}$ are the weighting coefficients of the corresponding loss terms. Here, $\mathcal{L}_{\mathrm{cls}}$ denotes the classification loss, $\mathcal{L}_{\mathrm{CIoU}}$ supervises bounding-box overlap and geometric consistency, and $\mathcal{L}_{\mathrm{DFL}}$ optimizes the discrete distributions used for box regression. The same objective is adopted for the principal architecture comparisons unless otherwise specified in the ablation studies.

\section{Experiments}
\label{sec:experiments}

This section evaluates HDMamba-YOLO on VisDrone2019 and AI-TOD from three complementary perspectives. First, we compare the proposed Tiny, Lite, and Base variants with representative detectors to characterize their accuracy--complexity trade-offs on VisDrone2019. Second, we evaluate HDMamba-YOLO-B on AI-TOD under a unified COCO-style protocol to examine its generalization to aerial scenes dominated by extremely small objects. Third, a set of controlled ablation studies is conducted on VisDrone2019 to analyze the contributions of the perception, reconstruction, alignment, and interaction stages, as well as the localization-loss configuration.

\subsection{Datasets and Evaluation Metrics}
\label{subsec:datasets_metrics}

\subsubsection{Datasets}
\textbf{VisDrone2019.}
VisDrone2019 is a UAV object detection benchmark collected under diverse urban and suburban conditions~\citep{du2019visdrone}. It contains scenes with dense object distributions, substantial scale variation, partial occlusion, and complex backgrounds. We use the official training and validation splits, consisting of 6,471 training images and 548 validation images, respectively. The dataset contains ten object categories: pedestrian, people, bicycle, car, van, truck, tricycle, awning-tricycle, bus, and motor. All VisDrone experiments in this work use the same 548-image validation split (denoted internally as \texttt{VisDrone\_full548}).

\textbf{AI-TOD.}
AI-TOD is designed for aerial tiny-object detection and contains a large proportion of objects represented by only a few pixels~\citep{wang2021tiny}. We follow the official split used in our experiments, with 11,214 training images and 2,804 validation images. The corresponding annotations contain 282,580 and 70,424 object instances, respectively, for a total of 353,004 annotated instances. The eight categories are airplane, bridge, storage-tank, ship, swimming-pool, vehicle, person, and wind-mill. No validation image is used for training. Table~\ref{tab:dataset_statistics} summarizes the dataset splits used in the experiments.

\begin{table}[!t]
\centering
\caption{Dataset splits used in the experiments.}
\label{tab:dataset_statistics}
\renewcommand{\arraystretch}{1.08}
\setlength{\tabcolsep}{7pt}
\footnotesize
\begin{tabular}{lccc}
\toprule
Dataset & Categories & Training images & Validation images \\
\midrule
VisDrone2019 & 10 & 6,471 & 548 \\
AI-TOD & 8 & 11,214 & 2,804 \\
\bottomrule
\end{tabular}
\end{table}

The VisDrone2019 training set is dominated by small-scale instances. Using the equivalent side length $s=\sqrt{w_{\rm px}h_{\rm px}}$, 89,264 objects (26.01\%) satisfy $s<16$ pixels, 118,340 objects (34.48\%) fall within $16\leq s<32$ pixels, and 135,600 objects (39.51\%) have $s\geq32$ pixels. Thus, 60.49\% of the training instances have an equivalent side length below 32 pixels, providing a suitable setting for evaluating the proposed small-object-oriented design.

\subsubsection{Evaluation Metrics}
For VisDrone2019, Precision ($P$), Recall ($R$), mAP$_{50}$, and mAP$_{50:95}$ are used in the controlled experiments. The main cross-method comparison focuses on mAP$_{50}$ and mAP$_{50:95}$ because these metrics are consistently available across the selected comparison methods. Model complexity is measured using the number of parameters and GFLOPs. For HDMamba-YOLO, the reported GFLOPs include an analytical compensation for Selective Scan operations that are not counted by the conventional profiling tool.

For AI-TOD, evaluation follows the COCO-style protocol implemented with \texttt{pycocotools.COCOeval}~\citep{lin2014microsoft}. We report AP, AP$_{50}$, AP$_{75}$, AP$_{vt}$, AP$_t$, AP$_s$, AR$_{vt}$, AR$_t$, and AR$_s$. The AI-TOD-specific area ranges are set to $[4,64]$, $[64,256]$, $[256,1024]$, and $[1024,4096]$ pixels$^2$ for very tiny (vt), tiny (t), small (s), and medium (m) objects, respectively. AP$_m$ and AR$_m$ are additionally reported in the text for HDMamba-YOLO-B.

\subsection{Experimental Environment and Implementation Details}
\label{subsec:implementation}

All experiments were conducted under Ubuntu 22.04.3 LTS on a workstation equipped with an Intel Xeon Platinum 8352V CPU and a single NVIDIA GeForce RTX 4090 GPU with 24 GB of memory. The models were implemented using Python 3.10.19, PyTorch 2.1.0, CUDA 12.1, and cuDNN 8.9.2, with the detection pipeline developed on Ultralytics 8.3.9.

All HDMamba-YOLO variants were trained from scratch without ImageNet or object-detection pretraining. For VisDrone2019, the input resolution was fixed to $640\times640$, the batch size was 8, and the maximum training length was 300 epochs. SGD was used with an initial learning rate of 0.01, momentum of 0.937, and weight decay of $5\times10^{-4}$. The learning rate followed a cosine schedule with a final ratio of 0.01, corresponding to a scheduled endpoint of $1\times10^{-4}$, and the first three epochs were used for warm-up. Automatic mixed-precision training was enabled, the random seed was fixed to 0, and early stopping was applied with a patience of 15 epochs. The loss coefficients for box regression, classification, and distribution focal loss were set to 7.5, 0.5, and 1.5, respectively, with CIoU used for box regression. Data augmentation included HSV perturbation $(0.015,0.7,0.4)$, translation of 0.1, scaling of 0.5, horizontal flipping with a probability of 0.5, and Mosaic augmentation with a probability of 1.0. MixUp and Copy-Paste were disabled, and Mosaic was closed during the final ten scheduled epochs.

For AI-TOD, the maximum training length was increased to 400 epochs, while the mini-batch size remained 8. A nominal batch size of 64 was used with gradient accumulation over eight mini-batches. Early-stopping patience was increased to 30 epochs, and a linear learning-rate schedule (\texttt{cos\_lr=False}) was adopted; the remaining core optimization settings were kept unchanged. All controlled ablations and models reproduced within each dataset followed the same data split, initialization policy, and training protocol. Results adopted from previously published studies retain their original experimental settings and are used only to characterize representative accuracy--complexity operating points.

\subsubsection{Unified AI-TOD Evaluation Protocol}
\label{subsubsec:aitod_protocol}
To minimize evaluation-side discrepancies, the reproduced AI-TOD models are reloaded from their respective best checkpoints and evaluated on the same official 2,804-image validation set. Inference is performed with an input size of 640, confidence threshold of 0.001, NMS IoU threshold of 0.7, and a maximum of 300 retained detections per image. Test-time augmentation is disabled and class-aware NMS is used. Predicted boxes are mapped back to absolute coordinates in the original image space and converted to COCO format using the official image and category identifiers.

The evaluator uses
\begin{equation}
\mathrm{maxDets}=[1,10,300],
\end{equation}
and the AI-TOD-specific area intervals described above. For HDMamba-YOLO-B, 225,576 predictions were successfully converted, with no prediction discarded because of an image-ID or category-ID mismatch. Re-generating the prediction JSON using the unified inference pipeline produced the same object-level predictions as the original official evaluation output. The final HDMamba-YOLO-B metrics are therefore reported from this unified COCO-style evaluation rather than from the internal Ultralytics validation summary.

\subsection{Results on VisDrone2019}
\label{subsec:visdrone_results}

\subsubsection{Comparison with Representative Detectors}
\label{subsubsec:visdrone_comparison}
Table~\ref{tab:visdrone_comparison} compares the proposed variants with representative detectors. To avoid conflating models with substantially different deployment budgets, the methods are organized into three parameter ranges. The proposed model is listed last within each range for readability. The best and second-best values within each parameter range are highlighted in bold and underlined, respectively.

\begin{table*}[!t]
\centering
\caption{Comparison with representative detectors on the VisDrone2019 validation set. Models are grouped according to parameter budgets. Within each group, the proposed HDMamba-YOLO variant is listed after the comparison methods. The best result is shown in bold and the second-best result is underlined.}
\label{tab:visdrone_comparison}
\renewcommand{\arraystretch}{1.08}
\setlength{\tabcolsep}{4.3pt}
\footnotesize
\begin{adjustbox}{max width=\textwidth}
\begin{tabular}{llcccc}
\toprule
Budget & Method & Params (M) $\downarrow$ & GFLOPs $\downarrow$ & mAP$_{50}$ (\%) $\uparrow$ & mAP$_{50:95}$ (\%) $\uparrow$ \\
\midrule
\multirow{5}{*}{$\leq3$M}
& YOLO11n & 2.59 & \underline{6.500} & 33.400 & 19.100 \\
& YOLOv10-n & 2.71 & 8.400 & 33.300 & 19.000 \\
& YOLOv8n & 3.00 & 8.100 & 34.100 & 19.500 \\
& SFBF-YOLO-n & \textbf{1.06} & 13.700 & \textbf{41.000} & \textbf{24.300} \\
& \textbf{HDMamba-YOLO-Tiny (Ours)} & \underline{1.34} & \textbf{6.419} & \underline{35.683} & \underline{21.097} \\
\midrule
\multirow{4}{*}{3--8M}
& SFBF-YOLO-s & \textbf{3.61} & 42.600 & \textbf{47.900} & \textbf{29.200} \\
& Mamba-YOLO-T & 5.80 & \textbf{13.200} & 37.867 & 22.715 \\
& LMF-UAV(s) & 6.30 & \underline{17.600} & 36.400 & 21.300 \\
& \textbf{HDMamba-YOLO-Lite (Ours)} & \underline{5.34} & 19.304 & \underline{41.140} & \underline{24.741} \\
\midrule
\multirow{8}{*}{8--20M}
& YOLOv10-s & \textbf{8.10} & 24.800 & 39.100 & 22.900 \\
& YOLOv12s & \underline{9.23} & \textbf{21.200} & 39.500 & 23.600 \\
& YOLO11s & 9.43 & \underline{21.600} & 39.600 & 23.200 \\
& YOLOv8s & 11.10 & 28.700 & 40.000 & 23.800 \\
& KSCNet & 12.10 & 33.000 & \underline{41.300} & \underline{24.700} \\
& Hyper-YOLO-S & 14.80 & 38.900 & 40.900 & 24.400 \\
& RT-DETR-R18 & 19.88 & 57.000 & 40.900 & 24.100 \\
& \textbf{HDMamba-YOLO-B (Ours)} & 10.04 & 29.879$^{\dagger}$ & \textbf{42.737} & \textbf{25.713} \\
\bottomrule
\end{tabular}
\end{adjustbox}
\vspace{1mm}
\begin{minipage}{0.98\textwidth}
\footnotesize
\textit{Sources:} YOLO11n and YOLO11s \citep{khanam2024yolov11}; YOLOv10-n and YOLOv10-s \citep{wang2024yolov10}; YOLOv8n and YOLOv8s \citep{jocher2023yolov8}; SFBF-YOLO-n and SFBF-YOLO-s \citep{hua2026sfbfyolo}; Mamba-YOLO-T \citep{wang2025mamba}; LMF-UAV(s) \citep{yang2025lmfuav}; YOLOv12s \citep{tian2025yolov12}; KSCNet \citep{li2025kscnet}; Hyper-YOLO-S \citep{feng2025hyperyolo}; and RT-DETR-R18 \citep{zhao2024rtdetr}.\par\smallskip
\textit{Note:} $^{\dagger}$ The GFLOPs of HDMamba-YOLO include the analytically compensated computational cost of Selective Scan operations. Cross-paper results are used to characterize representative accuracy--complexity operating points; reproduced methods and all HDMamba-YOLO variants follow our unified experimental setting where applicable.
\end{minipage}
\end{table*}

The three HDMamba-YOLO variants provide distinct operating points. HDMamba-YOLO-Tiny uses only 1.342M parameters and 6.419 GFLOPs, giving the lowest computational cost in the ultra-lightweight group. HDMamba-YOLO-Lite increases the capacity to 5.344M parameters and achieves 41.140\% mAP$_{50}$ and 24.741\% mAP$_{50:95}$, providing an intermediate point between the Tiny and Base variants. HDMamba-YOLO-B reaches 42.737\% mAP$_{50}$ and 25.713\% mAP$_{50:95}$ with 10.042M parameters and 29.879 corrected GFLOPs.

Within the compact group, KSCNet provides a strong reference with 41.300\% mAP$_{50}$ and 24.700\% mAP$_{50:95}$. HDMamba-YOLO-B improves these two metrics by 1.437 and 1.013 percentage points, respectively, while using fewer parameters (10.042M versus 12.10M) and lower reported computation (29.879 versus 33.000 GFLOPs). This comparison indicates that the proposed Base configuration occupies a favorable accuracy--complexity operating point within the evaluated compact-model range.

In the lightweight group, SFBF-YOLO-s attains the highest detection accuracy, whereas Mamba-YOLO-T has the lowest reported computational cost. HDMamba-YOLO-Lite lies between these operating points. Compared with the reproduced Mamba-YOLO-T at a similar parameter scale, HDMamba-YOLO-Lite improves mAP$_{50}$ and mAP$_{50:95}$ by 3.273 and 2.026 percentage points, respectively, while requiring higher computation (19.304 versus 13.200 GFLOPs). Under the present VisDrone setting, this comparison shows that the stage-wise allocation used by HDMamba-YOLO-Lite reaches a higher-accuracy operating point while requiring moderately greater computation than Mamba-YOLO-T.

\begin{figure*}[!t]
    \centering
    \includegraphics[
        width=0.98\textwidth,
        keepaspectratio
    ]{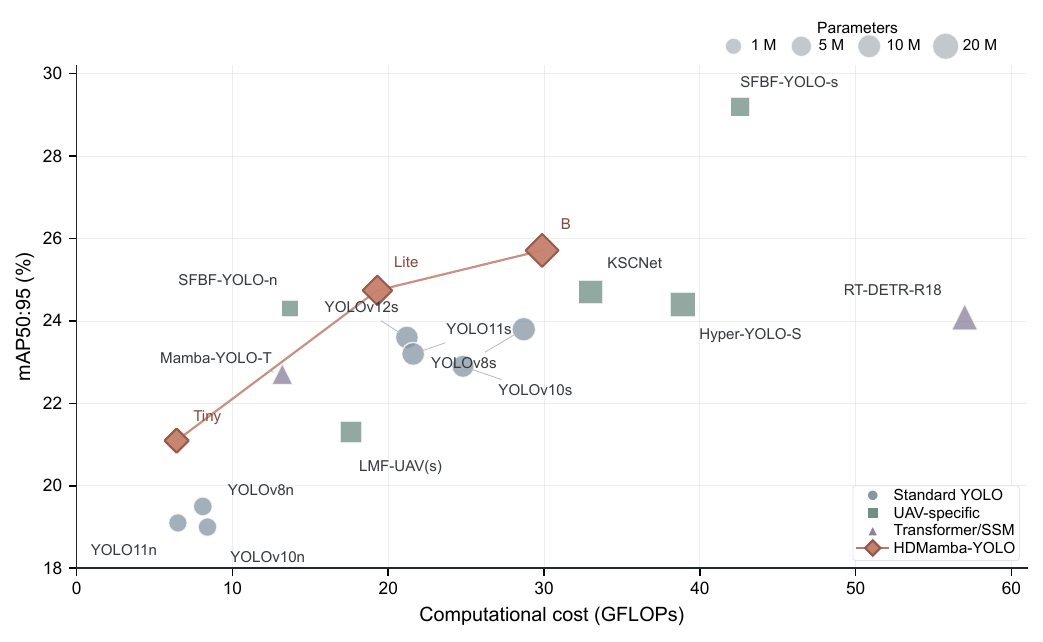}
    \caption{Accuracy--complexity comparison on the VisDrone2019 validation set. The horizontal axis denotes reported or corrected GFLOPs, the vertical axis denotes mAP$_{50:95}$, and marker area is proportional to the number of parameters. Marker shapes distinguish detector families, while the connected diamond markers indicate the Tiny, Lite, and Base configurations of HDMamba-YOLO.}
    \label{fig:visdrone_accuracy_complexity}
\end{figure*}

Figure~\ref{fig:visdrone_accuracy_complexity} complements Table~\ref{tab:visdrone_comparison} by visualizing the accuracy--complexity operating points of the compared detectors. Computational cost is shown on the horizontal axis, mAP$_{50:95}$ on the vertical axis, and marker area is proportional to the number of parameters. The three HDMamba-YOLO variants are connected only to indicate their capacity progression from Tiny to Lite to Base. The trajectory shows a consistent increase in detection accuracy as capacity grows, while retaining substantially lower computational cost than several higher-capacity comparison methods. The resulting distribution provides a direct visual summary of the different accuracy--complexity operating points represented in Table~\ref{tab:visdrone_comparison}.

\subsubsection{Qualitative Analysis on VisDrone2019}
Quantitative results are complemented by feature-response and detection visualizations on three representative validation scenes containing dense traffic participants, cluttered road backgrounds, partial occlusion, and distant small instances. YOLO11s is used as the primary reference because it has a parameter scale comparable to HDMamba-YOLO-B and is also included in the quantitative comparison.

\begin{figure*}[!t]
    \centering
    \includegraphics[
        width=0.98\textwidth,
        keepaspectratio
    ]{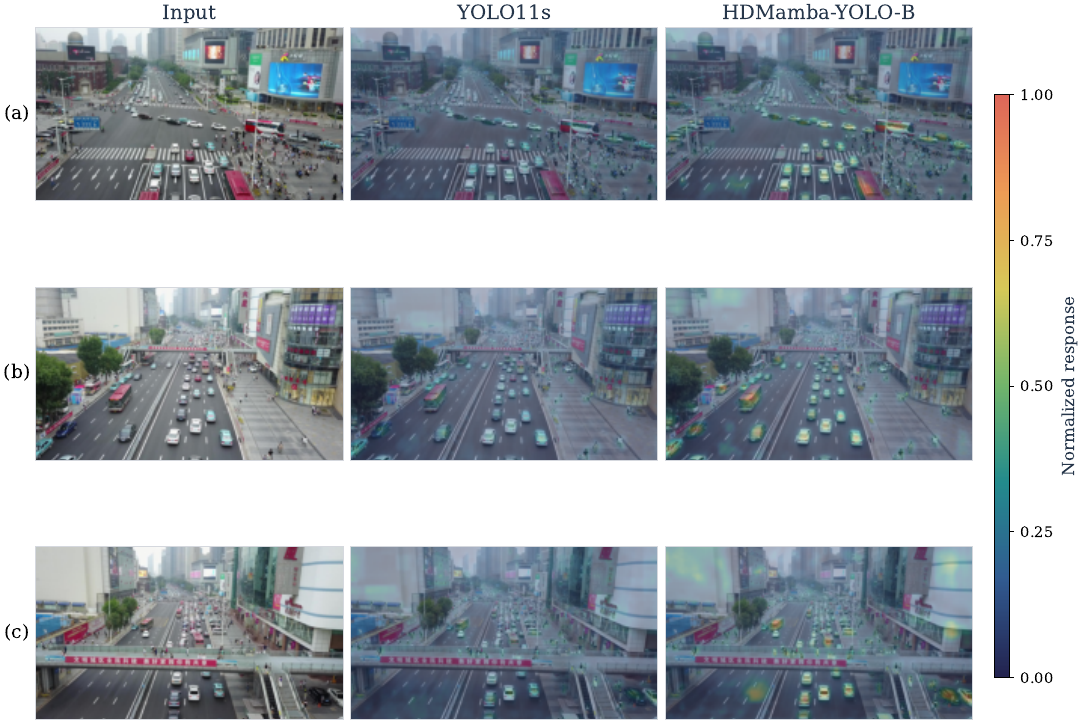}
    \caption{Feature-response comparison on representative VisDrone2019 validation scenes. Rows (a)--(c) represent dense mixed traffic and pedestrian clusters, low-contrast vehicles against cluttered road backgrounds, and densely packed distant targets with partial occlusion, respectively. Columns show the input image, YOLO11s response, and HDMamba-YOLO-B response, respectively. Responses are extracted from the P3 pre-detection feature (stride 8) and summarized by channel-wise root-mean-square activation. Within each row, the two models use the same normalization scale; warmer colors indicate larger normalized responses.}
    \label{fig:visdrone_feature_response}
\end{figure*}

Figure~\ref{fig:visdrone_feature_response} compares the spatial response patterns of the two detectors at the P3 pre-detection feature level (stride 8). For each image, the response map is computed from the channel-wise root-mean-square activation and the two models are normalized using a shared image-specific scale. Consequently, color intensity can be compared between YOLO11s and HDMamba-YOLO-B within the same row, while absolute colors should not be compared across different scenes. Across the three examples, HDMamba-YOLO-B exhibits more spatially concentrated responses around several dense vehicle and pedestrian regions, whereas the YOLO11s responses are more diffuse in parts of the road and background. These visualizations are consistent with stronger target--background discrimination in the proposed representation, but they are used only as qualitative evidence and do not by themselves establish the causal contribution of any individual module.

\begin{figure*}[!t]
    \centering
    \includegraphics[
        width=0.98\textwidth,
        keepaspectratio
    ]{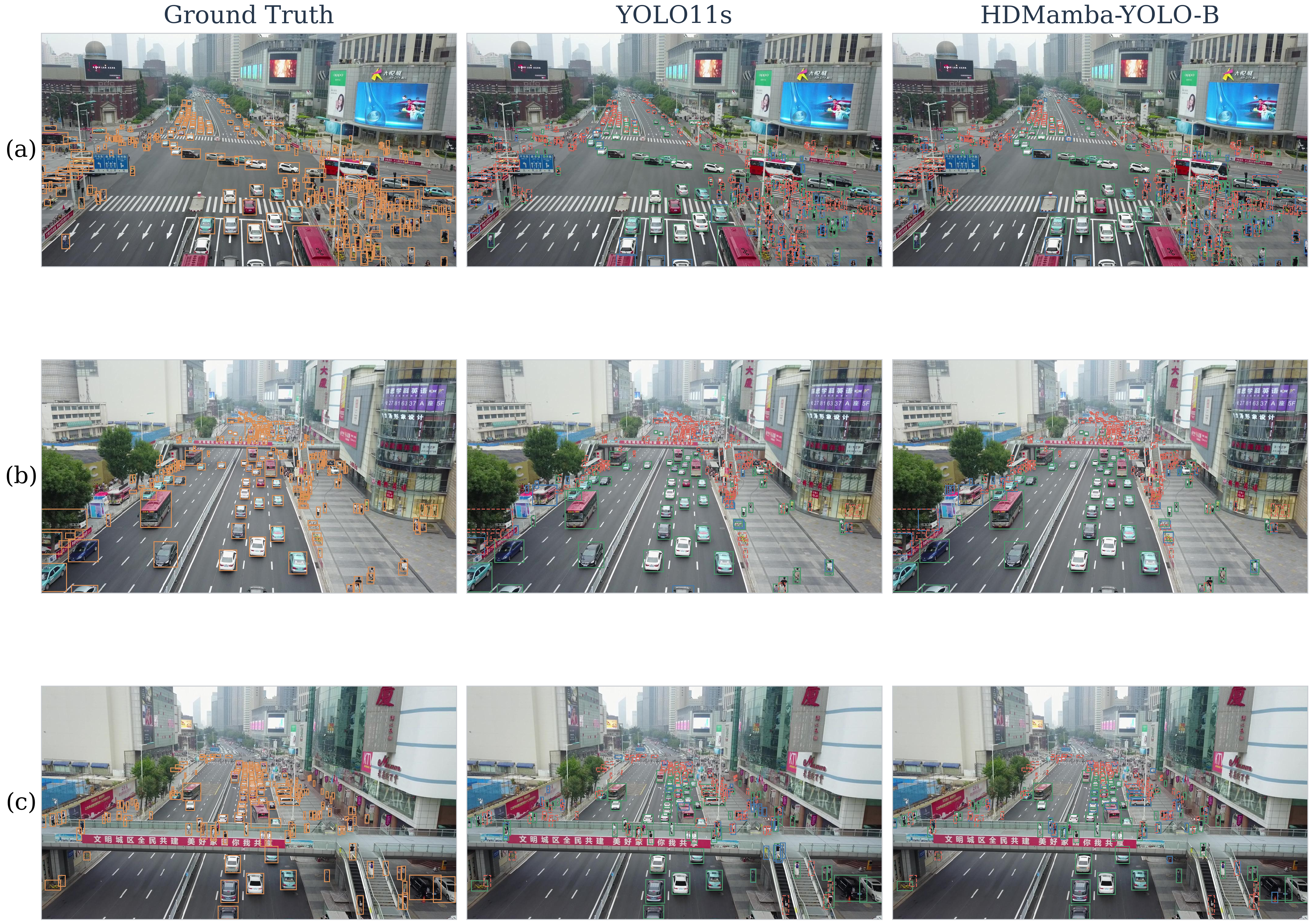}
    \caption{Qualitative detection comparison on representative VisDrone2019 validation scenes. The three rows follow the same scene order and challenge categories as Fig.~\ref{fig:visdrone_feature_response}, enabling a direct comparison between feature-response patterns and the corresponding detection outcomes. Columns show the ground-truth annotations, YOLO11s predictions, and HDMamba-YOLO-B predictions, respectively. Orange boxes indicate ground-truth instances. In the prediction panels, green and blue solid boxes denote true-positive (TP) and false-positive (FP) detections, respectively, while red dashed boxes indicate unmatched ground-truth objects (FN). Visual matching uses class-consistent one-to-one assignment at an IoU threshold of 0.5; detections are shown at a confidence threshold of 0.25 with an NMS IoU threshold of 0.7.} 
    \label{fig:visdrone_detection}
\end{figure*}

Figure~\ref{fig:visdrone_detection} further compares the final detections on the same scene set under identical inference settings. The three columns present the ground-truth annotations, YOLO11s predictions, and HDMamba-YOLO-B predictions in full-scene views, allowing the corresponding TP, FP, and FN patterns to be compared directly. In dense and partially occluded regions, HDMamba-YOLO-B recovers additional valid instances in several clusters while maintaining a comparable overall detection pattern to YOLO11s in less challenging regions. These selected examples provide qualitative evidence that is consistent with the quantitative improvement in mAP reported above.

\subsection{Results on AI-TOD}
\label{subsec:aitod_results}

\subsubsection{Quantitative Comparison}
Table~\ref{tab:aitod_comparison} reports the AI-TOD results. For the reproduced models, the final metrics are obtained using the unified inference and COCO-style evaluation protocol described in Section~\ref{subsubsec:aitod_protocol}. Methods whose results are taken from the literature are included only when their reported metric definitions are compatible with the AI-TOD evaluation columns used here. The purpose of this table is to compare representative operating points rather than to claim universal superiority across all detector families.

\begin{table*}[!t]
\centering
\caption{Comparison with representative detectors on the AI-TOD validation set. The best and second-best values are highlighted in bold and underlined, respectively.}
\label{tab:aitod_comparison}
\renewcommand{\arraystretch}{1.08}
\setlength{\tabcolsep}{2.4pt}
\scriptsize
\begin{adjustbox}{max width=\textwidth}
\begin{tabular}{lcccccccccccc}
\toprule
Method & Params (M) $\downarrow$ & Resolution & AP $\uparrow$ & AP$_{50}$ $\uparrow$ & AP$_{75}$ $\uparrow$ & AP$_{vt}$ $\uparrow$ & AP$_t$ $\uparrow$ & AP$_s$ $\uparrow$ & AR$_{vt}$ $\uparrow$ & AR$_t$ $\uparrow$ & AR$_s$ $\uparrow$ & GFLOPs $\downarrow$ \\
\midrule
ALSS-YOLO-S & \textbf{2.18} & $640\times640$ & 12.900 & 31.770 & 7.550 & 2.220 & 12.830 & 19.940 & 3.670 & 22.980 & 30.450 & \textbf{8.50} \\
ALSS-YOLO-M & \underline{2.74} & $640\times640$ & 12.940 & 32.830 & 7.830 & 2.400 & 12.420 & 20.200 & 4.920 & 22.420 & 31.330 & \underline{10.40} \\
YOLOv9-S & 7.29 & $640\times640$ & 20.330 & 47.540 & 14.160 & 7.520 & 20.130 & 27.960 & 13.650 & 33.680 & 39.170 & 27.08 \\
YOLOv10s & 8.07 & $640\times640$ & 19.780 & 45.540 & 13.950 & 6.640 & 20.380 & 26.970 & \underline{15.090} & \underline{35.890} & 40.430 & 24.80 \\
YOLO11s & 9.43 & $640\times640$ & 20.250 & 46.870 & 14.660 & \underline{9.030} & 20.510 & 27.780 & 13.340 & 33.490 & 40.450 & 21.60 \\
YOLOv8-S & 11.14 & $640\times640$ & 20.800 & \underline{48.450} & 14.690 & 5.730 & 20.970 & 29.100 & 9.560 & 34.410 & 40.650 & 28.45 \\
YOLOv8m & 25.86 & $640\times640$ & \textbf{23.510} & \textbf{52.680} & \textbf{18.530} & \textbf{13.300} & \textbf{23.460} & \textbf{31.240} & \textbf{20.590} & \textbf{38.050} & \textbf{42.380} & 78.70 \\
\textbf{HDMamba-YOLO-B (Ours)} & 10.04 & $640\times640$ & \underline{21.621} & 47.881 & \underline{15.986} & 6.333 & \underline{22.167} & \underline{29.718} & 12.213 & 35.355 & \underline{41.638} & 29.88$^{\dagger}$ \\
\bottomrule
\end{tabular}
\end{adjustbox}
\vspace{1mm}
\begin{minipage}{0.98\textwidth}
\scriptsize
\textit{Sources:} ALSS-YOLO-S and ALSS-YOLO-M \citep{he2024alssyolo}; YOLOv9-S \citep{wang2024yolov9}; YOLOv10s \citep{wang2024yolov10}; YOLO11s \citep{khanam2024yolov11}; and YOLOv8-S and YOLOv8m \citep{jocher2023yolov8}.\par\smallskip
\textit{Note:} $^{\dagger}$ The GFLOPs of HDMamba-YOLO-B include Selective Scan compensation. HDMamba-YOLO-B additionally obtains AP$_m=36.826\%$ and AR$_m=42.935\%$ for the 32--64-pixel scale range.
\end{minipage}
\end{table*}

HDMamba-YOLO-B obtains 21.621\% AP, 47.881\% AP$_{50}$, and 15.986\% AP$_{75}$. Relative to YOLO11s, which has a similar parameter scale, the proposed model improves AP, AP$_{50}$, and AP$_{75}$ by 1.371, 1.011, and 1.326 percentage points, respectively. The gains are also observed for the tiny and small ranges: AP$_t$ increases from 20.510\% to 22.167\%, AP$_s$ from 27.780\% to 29.718\%, AR$_t$ from 33.490\% to 35.355\%, and AR$_s$ from 40.450\% to 41.638\%. These results indicate that the improvement extends beyond a single IoU threshold and is reflected in both precision and recall for the 8--32-pixel scale ranges.

YOLOv8m achieves the highest overall performance in Table~\ref{tab:aitod_comparison}, providing a useful higher-capacity reference. It uses approximately 2.58$\times$ the parameters and 2.63$\times$ the reported GFLOPs of HDMamba-YOLO-B. The AP gap between the two models is 1.889 percentage points, while the AP$_t$, AP$_s$, and AR$_s$ gaps are 1.293, 1.522, and 0.742 percentage points, respectively. The comparison therefore places HDMamba-YOLO-B at a lower-complexity operating point, while YOLOv8m retains higher overall accuracy.

For very tiny objects, HDMamba-YOLO-B obtains 6.333\% AP$_{vt}$ and 12.213\% AR$_{vt}$, which are lower than the corresponding YOLO11s values of 9.030\% and 13.340\%. This result identifies a remaining limitation: objects with an equivalent side length of only 2--8 pixels provide extremely limited visual evidence, and the current architecture does not improve this regime uniformly. We therefore interpret the AI-TOD results as evidence of improved performance primarily in the 8--32-pixel ranges rather than as an all-scale improvement.

\subsubsection{Scale-Aware Qualitative Analysis}
\begin{figure*}[!t]
    \centering
    \includegraphics[
        width=0.98\textwidth,
        keepaspectratio
    ]{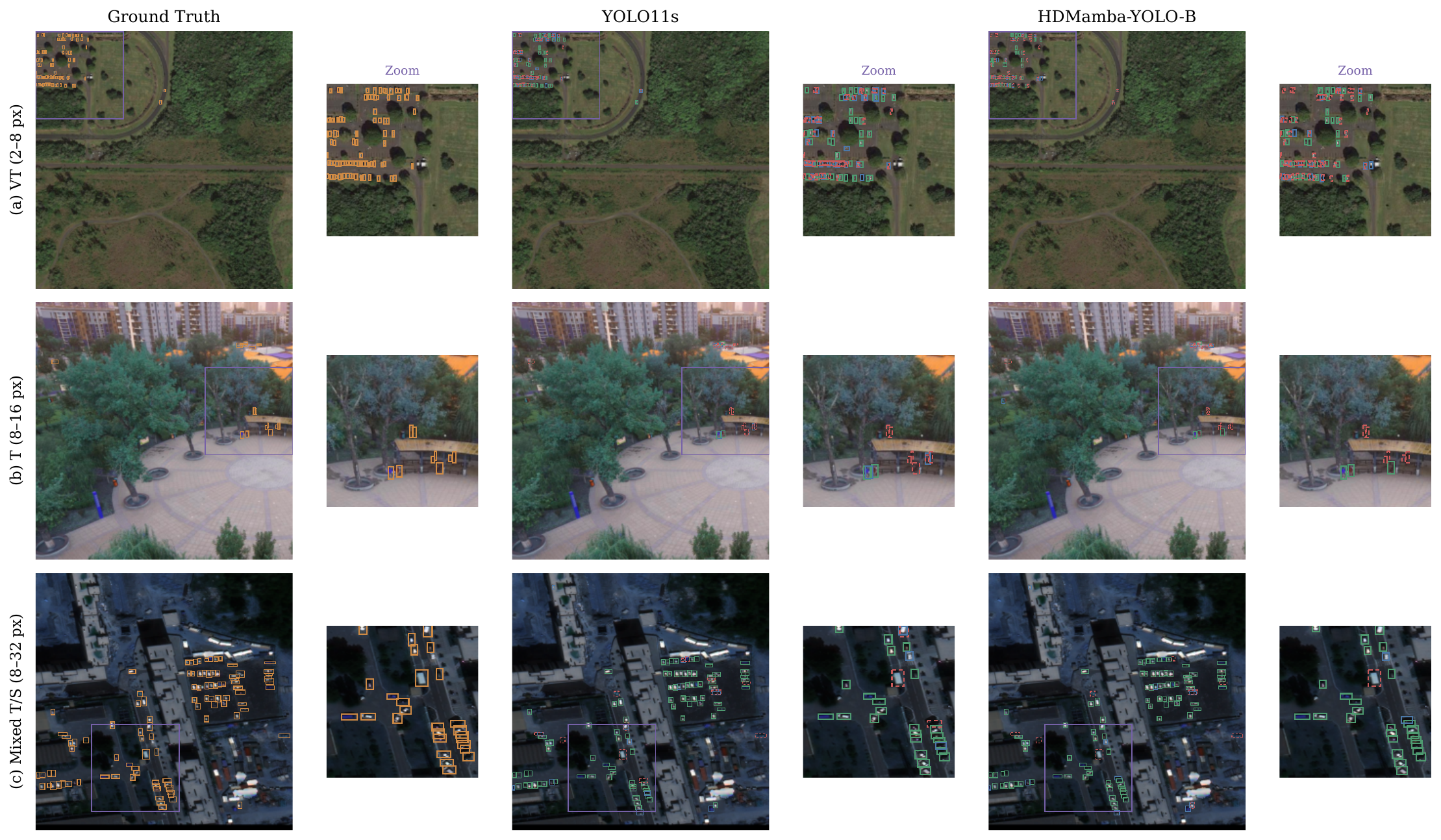}
    \caption{Scale-aware qualitative detection comparison on the official AI-TOD validation set. Rows (a)--(c) correspond to a very-tiny-dominated scene (image 1697), a tiny-object scene (image 1043), and a mixed tiny/small scene (image 1845), respectively. Columns show Ground Truth, YOLO11s, and HDMamba-YOLO-B. Orange boxes denote ground-truth instances; green and blue solid boxes in prediction panels denote TP and FP detections, while red dashed boxes denote unmatched ground-truth objects (FN). Purple rectangles indicate the ground-truth-driven regions shown in the enlarged views, and the same region is used across the three columns of each row.}
    \label{fig:aitod_scale_aware_detection}
\end{figure*}

Figure~\ref{fig:aitod_scale_aware_detection} complements the scale-specific AP/AR results with three validation scenes selected from ground-truth-only candidate pools before model comparison. Row (a) is dominated by very-tiny objects (image 1697; 86 vt and 1 t instances), row (b) contains only tiny objects (image 1043; 16 t instances), and row (c) is a mixed tiny/small scene (image 1845; 47 t and 53 s instances). For each scene, the same ground-truth-driven region of interest is used for YOLO11s and HDMamba-YOLO-B.

The scene-level TP/FP/FN counts reveal scale-dependent behavior that is consistent with the quantitative evaluation. In the very-tiny scene, YOLO11s records 36/31/51 TP/FP/FN, whereas HDMamba-YOLO-B records 30/16/57. The proposed model therefore produces fewer false positives but also fewer true positives and more misses, which is consistent with the lower AP$_{vt}$ and AR$_{vt}$ reported in Table~\ref{tab:aitod_comparison}. In the tiny-object scene, the counts change from 3/3/13 to 4/3/12, corresponding to a modest improvement. In the mixed tiny/small scene, they change from 88/13/12 to 91/11/9, showing a clearer increase in recovered objects together with fewer false positives and misses. These counts describe the selected scenes only and should not be interpreted as dataset-level precision or recall.

\subsection{Ablation Studies}
\label{subsec:ablation}

All ablation studies are conducted on \texttt{VisDrone\_full548}. Unless explicitly stated, the compared variants use the same input resolution, batch size, optimizer family, seed, pure-CIoU objective, augmentation pipeline, and early-stopping criterion. The best checkpoint of each run is selected using the Ultralytics validation fitness, $0.1\times\mathrm{mAP}_{50}+0.9\times\mathrm{mAP}_{50:95}$. The ablation sequence follows the functional design of HDMamba-YOLO: perception, reconstruction, alignment, and interaction.

\subsubsection{Progressive Component Ablation}
\label{subsubsec:progressive_ablation}
Table~\ref{tab:component_ablation} first evaluates the progressive construction of the complete model. Starting from the YOLO11s baseline, EVSS, DST-Wrapper, Native C3k2-ASSAF, DySample, and OS-CVTIA are introduced successively. This table is intended to quantify the incremental effect of the complete processing pipeline; more targeted controls for individual design choices are provided in the subsequent subsections.

\begin{table*}[!t]
\centering
\caption{Progressive component ablation of HDMamba-YOLO on VisDrone2019.}
\label{tab:component_ablation}
\renewcommand{\arraystretch}{1.08}
\setlength{\tabcolsep}{3.1pt}
\footnotesize
\begin{adjustbox}{max width=\textwidth}
\begin{tabular}{lccccccccccc}
\toprule
Config. & EVSS & DST & C3k2-ASSAF & DySample & OS-CVTIA &
P (\%) & R (\%) & mAP$_{50}$ (\%) & mAP$_{50:95}$ (\%) &
Params (M) $\downarrow$ & GFLOPs $\downarrow$ \\
\midrule
A0 & -- & -- & -- & -- & -- &
49.414 & 38.910 & 39.400 & 23.735 & 9.432 & \textbf{21.568} \\

A1 & \checkmark & -- & -- & -- & -- &
52.035 & 40.470 & 41.613 & 25.173 & \textbf{9.210} & 28.877 \\

A2 & \checkmark & \checkmark & -- & -- & -- &
51.649 & 40.561 & 41.792 & 25.265 & 9.749 & 29.204 \\

A3 & \checkmark & \checkmark & \checkmark & -- & -- &
53.006 & 40.972 & 42.431 & 25.542 & 9.763 & 29.166 \\

A4 & \checkmark & \checkmark & \checkmark & \checkmark & -- &
\textbf{53.747} & 40.525 & 42.388 & 25.569 & 9.773 & 29.182 \\

\textbf{A5} & \checkmark & \checkmark & \checkmark & \checkmark & \checkmark &
53.367 & \textbf{41.572} & \textbf{42.737} & \textbf{25.713} &
10.042 & 29.879 \\
\bottomrule
\end{tabular}
\end{adjustbox}
\end{table*}

The largest single increase occurs when EVSS is introduced in A1: relative to A0, mAP$_{50}$ and mAP$_{50:95}$ increase by 2.213 and 1.438 percentage points, respectively. This result supports the use of state-space modeling for the backbone-level perception stage under the present UAV setting. Adding DST-Wrapper in A2 provides smaller but positive gains of 0.179 and 0.092 percentage points in mAP$_{50}$ and mAP$_{50:95}$, respectively, indicating that an explicit transition between backbone perception and neck reconstruction is beneficial in the final pipeline.

Introducing Native C3k2-ASSAF in A3 increases mAP$_{50}$ from 41.792\% to 42.431\% and mAP$_{50:95}$ from 25.265\% to 25.542\%. DySample in A4 produces a nearly unchanged mAP$_{50}$ (42.388\%) and a small increase in mAP$_{50:95}$ (25.569\%); its role is therefore examined more carefully using a two-condition control in Section~\ref{subsubsec:dysample_ablation}. Finally, replacing the native head with OS-CVTIA increases recall from 40.525\% to 41.572\%, mAP$_{50}$ from 42.388\% to 42.737\%, and mAP$_{50:95}$ from 25.569\% to 25.713\%, while precision decreases slightly from 53.747\% to 53.367\%. The progressive study thus supports the complete pipeline without requiring every component to improve every metric independently.

\subsubsection{Stage-wise Allocation of SSM and CNN Operators}
\label{subsubsec:ssm_cnn}
The progressive study does not by itself isolate whether the final performance is caused by the stage-wise allocation of SSM and CNN operators. We therefore construct two complementary controls around A5. CNN Full-Control replaces the EVSS perception stages with channel- and stride-matched CNN/InRes processing while retaining PhasePatchMerging2D, DST-Wrapper, Native C3k2-ASSAF, DySample, OS-CVTIA, detection scales, and the training protocol. In the opposite direction, SSM-dominant Neck retains the EVSS backbone but replaces Native C3k2-ASSAF in the neck with backbone-homologous EVSS-based processing. These controls are designed to compare architectural allocation rather than to make a general statement about CNNs or SSMs as model families. The corresponding results are reported in Table~\ref{tab:ssm_cnn_paradigm_ablation}.

\begin{table*}[!t]
\centering
\caption{Controlled comparison of stage-wise SSM--CNN allocation on VisDrone2019.}
\label{tab:ssm_cnn_paradigm_ablation}
\renewcommand{\arraystretch}{1.08}
\setlength{\tabcolsep}{3.0pt}
\footnotesize
\begin{adjustbox}{max width=\textwidth}
\begin{tabular}{lcccccccccc}
\toprule
Configuration & Perception & Reconstruction & Alignment & Interaction &
P (\%) & R (\%) & mAP$_{50}$ (\%) & mAP$_{50:95}$ (\%) &
Params (M) $\downarrow$ & GFLOPs $\downarrow$ \\
\midrule
CNN Full-Control &
CNN/InRes & DST + Native ASSAF & DySample & OS-CVTIA &
50.743 & 38.426 & 39.824 & 23.843 &
\textbf{9.718} & \textbf{22.789} \\

SSM-dominant Neck &
EVSS & DST + EVSS & DySample & OS-CVTIA &
52.630 & 40.980 & 42.220 & 25.460 &
10.273 & 28.700 \\

\textbf{HDMamba-YOLO-B} &
\textbf{EVSS} & \textbf{DST + Native ASSAF} &
\textbf{DySample} & \textbf{OS-CVTIA} &
\textbf{53.367} & \textbf{41.572} &
\textbf{42.737} & \textbf{25.713} &
10.042 & 29.879 \\
\bottomrule
\end{tabular}
\end{adjustbox}
\end{table*}

Relative to CNN Full-Control, HDMamba-YOLO-B improves precision, recall, mAP$_{50}$, and mAP$_{50:95}$ by 2.624, 3.146, 2.913, and 1.870 percentage points, respectively. In the opposite control, replacing the CNN-style reconstruction operators in the neck with EVSS-based processing reduces the four metrics by 0.737, 0.592, 0.517, and 0.253 percentage points relative to A5. The latter difference is substantially smaller than the CNN-backbone control, but it is consistent across all four detection metrics.

These complementary observations support the proposed stage-wise allocation under the present controlled setting: state-space modeling is effective when assigned to backbone-level contextual perception, while explicit convolutional two-dimensional processing remains beneficial for repeated feature reconstruction and aggregation in the neck. The same tendency is consistent with the external Mamba-YOLO-T comparison in Table~\ref{tab:visdrone_comparison}, where the hybrid allocation used here reaches a higher-accuracy operating point under the evaluated UAV setting.

\subsubsection{Ablation of Backbone Stage-Transition Operators}
\label{subsubsec:stage_transition}
Table~\ref{tab:stage_transition_ablation} compares EfficientDownsample, LightweightPhaseDownsample, and PhasePatchMerging2D to isolate the choice of transition operator between high-resolution backbone stages under the same optimizer grouping and downstream architecture. The EfficientDownsample control was retrained using the same optimizer configuration as the two phase-based variants, so that the transition operator is the only intended structural variable at the corresponding backbone locations.

\begin{table*}[!t]
\centering
\caption{Ablation of backbone stage-transition operators on VisDrone2019.}
\label{tab:stage_transition_ablation}
\renewcommand{\arraystretch}{1.08}
\setlength{\tabcolsep}{4.5pt}
\footnotesize
\begin{adjustbox}{max width=\textwidth}
\begin{tabular}{lcccccc}
\toprule
Stage transition & P (\%) & R (\%) & mAP$_{50}$ (\%) & mAP$_{50:95}$ (\%) & Params (M) & GFLOPs \\
\midrule
EfficientDownsample & 51.963 & 40.611 & 41.678 & 25.292 & \textbf{9.921} & \textbf{29.256} \\
LightweightPhaseDownsample & 52.090 & 40.472 & 41.633 & 25.239 & 10.001 & 29.670 \\
\textbf{PhasePatchMerging2D} & \textbf{53.367} & \textbf{41.572} & \textbf{42.737} & \textbf{25.713} & 10.042 & 29.879 \\
\bottomrule
\end{tabular}
\end{adjustbox}
\end{table*}

Compared with EfficientDownsample, PhasePatchMerging2D improves precision, recall, mAP$_{50}$, and mAP$_{50:95}$ by 1.404, 0.961, 1.059, and 0.421 percentage points, respectively, with increases of 0.121M parameters and 0.623 GFLOPs. Compared with LightweightPhaseDownsample, the corresponding improvements are 1.277, 1.100, 1.104, and 0.474 percentage points, while the complexity difference is only 0.041M parameters and 0.209 GFLOPs.

Both phase-based operators use the same four spatial phases, whereas the final PhasePatchMerging2D concatenates $X_{00}$, $X_{10}$, $X_{01}$, and $X_{11}$, applies LayerNorm to the $4C$ representation, and uses a bias-free linear projection. The result therefore does not imply that four-phase sampling alone is sufficient; rather, the complete phase-aware merging formulation provides the best performance among the evaluated transition designs under the present protocol.

\subsubsection{Internal Reconstruction Analysis of C3k2-ASSAF}
\label{subsubsec:assaf_ablation}
A targeted branch ablation is conducted to clarify the roles of the two reconstruction streams in the final Native C3k2-ASSAF block. All variants use the same A3 outer topology, in which the input is split into two initial features, the second feature is iteratively processed by ASSAF, and the retained raw features and reconstructed response are concatenated before the final $1\times1$ projection. Only the internal ASSAF branch composition is changed. This design avoids mixing the present analysis with earlier prototype topologies or gate-search experiments.

\begin{table*}[!t]
\centering
\caption{Internal branch ablation of Native C3k2-ASSAF on VisDrone2019.}
\label{tab:assaf_branch_ablation}
\renewcommand{\arraystretch}{1.08}
\setlength{\tabcolsep}{4.5pt}
\footnotesize
\begin{adjustbox}{max width=\textwidth}
\begin{tabular}{lcccccccc}
\toprule
Variant & Local branch & Context branch & P (\%) & R (\%) & mAP$_{50}$ (\%) & mAP$_{50:95}$ (\%) & Params (M) & GFLOPs \\
\midrule
Local-only & \checkmark & -- & 52.512 & 40.771 & 42.116 & \textbf{25.545} & 9.763 & 29.116 \\
Context-only & -- & \checkmark & 50.873 & \textbf{41.790} & 42.001 & 25.240 & 9.763 & \textbf{28.877} \\
\textbf{Full C3k2-ASSAF} & \checkmark & \checkmark & \textbf{53.006} & 40.972 & \textbf{42.431} & 25.542 & 9.763 & 29.166 \\
\bottomrule
\end{tabular}
\end{adjustbox}
\end{table*}

The Local-only configuration achieves 52.512\% precision, 40.771\% recall, 42.116\% mAP$_{50}$, and 25.545\% mAP$_{50:95}$. The Context-only configuration yields a higher recall of 41.790\% but lower precision and AP. Relative to Context-only, Local-only changes precision, recall, and mAP$_{50:95}$ by $+1.639$, $-1.019$, and $+0.305$ percentage points, respectively. These results reveal different functional tendencies of the two branches: the orthogonal $1\times3\rightarrow3\times1$ branch is more closely associated with discriminative local reconstruction and localization quality, whereas the $7\times7$ depthwise context branch provides complementary neighborhood information and tends to favor recall.

Combining both branches in the full C3k2-ASSAF increases precision, recall, and mAP$_{50}$ over Local-only by 0.494, 0.201, and 0.315 percentage points, respectively. Its mAP$_{50:95}$ (25.542\%) is essentially identical to that of Local-only (25.545\%), with a difference of only 0.003 percentage points. Compared with Context-only, the full configuration improves precision, mAP$_{50}$, and mAP$_{50:95}$ by 2.133, 0.430, and 0.302 percentage points, respectively. We therefore interpret the full design as providing a more favorable overall precision--recall balance rather than claiming that every metric is maximized by the dual-branch configuration.

The adaptive independent-sigmoid gating used in the final implementation is retained as a lightweight modulation mechanism. The present controlled analysis, however, is centered on the more fundamental Local/Context branch complementarity; the observed gain is therefore not attributed solely to the gating operation.

\begin{figure*}[!t]
    \centering
    \includegraphics[
        width=0.98\textwidth,
        keepaspectratio
    ]{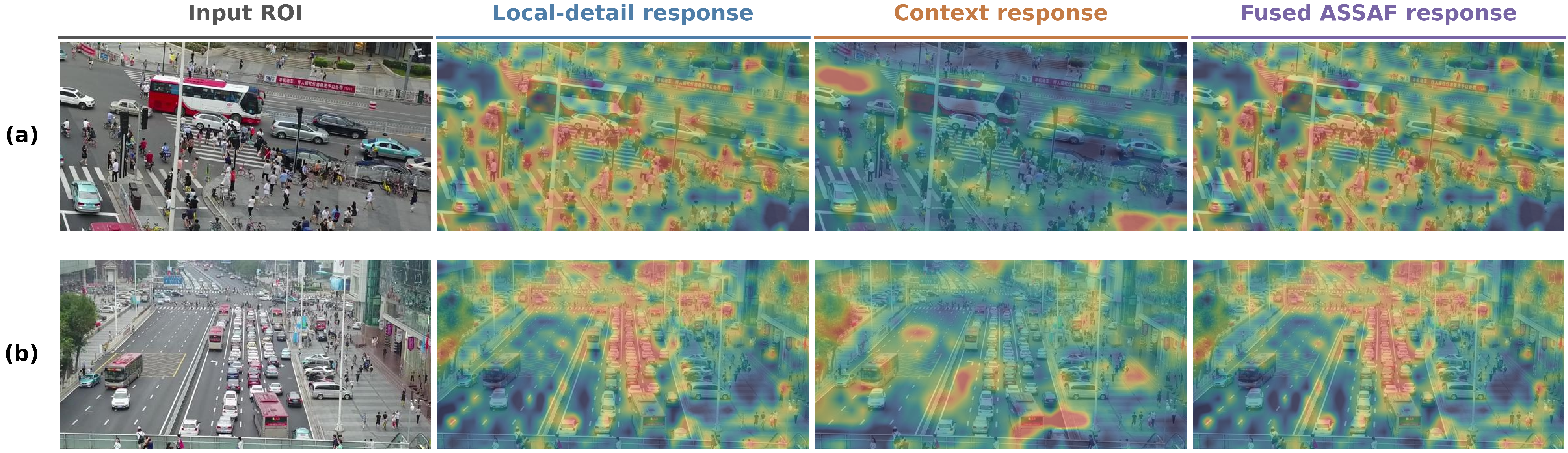}
    \caption{Internal response patterns of the full ASSAF block on two representative VisDrone regions. Columns show the input region, Local-detail response, Context response, and fused ASSAF response. The three response maps are extracted from the same forward pass. For visualization, each response is independently normalized using its 1st--99th percentile range; therefore, color intensity reflects the within-branch spatial distribution rather than absolute activation magnitude across branches.}
    \label{fig:assaf_local_context_response}
\end{figure*}

To examine the internal spatial tendencies behind these branch-level metrics, Fig.~\ref{fig:assaf_local_context_response} visualizes the Local-detail, Context, and fused ASSAF responses on two representative VisDrone regions. All three response maps are taken from the same forward pass of the full ASSAF block rather than from three separately trained models. The Local-detail response presents more concentrated spatial peaks, whereas the Context response provides broader and smoother support over neighboring structures. The fused response preserves the dominant localized pattern while incorporating lower-amplitude contextual support, which is consistent with the complementary behavior observed in Table~\ref{tab:assaf_branch_ablation}. Because each response is independently normalized using the 1st--99th percentile range, color intensity describes the within-branch spatial distribution; absolute activation magnitudes are therefore not comparable across branches.

\subsubsection{Controlled Ablation of DySample}
\label{subsubsec:dysample_ablation}
The progressive A3$\rightarrow$A4 comparison shows only a small change after DySample is introduced, making a single pair insufficient to characterize its role. We therefore evaluate DySample under both the native detection head and OS-CVTIA, as reported in Table~\ref{tab:dysample_ablation}. This two-condition design distinguishes the independent effect of dynamic upsampling from its interaction with the final head.

\begin{table*}[!t]
\centering
\caption{Controlled ablation of DySample under the native head and OS-CVTIA on VisDrone2019.}
\label{tab:dysample_ablation}
\renewcommand{\arraystretch}{1.08}
\setlength{\tabcolsep}{4.0pt}
\footnotesize
\begin{adjustbox}{max width=\textwidth}
\begin{tabular}{lcccccccc}
\toprule
Configuration & Upsampling & Head & P (\%) & R (\%) & mAP$_{50}$ (\%) & mAP$_{50:95}$ (\%) & Params (M) & GFLOPs \\
\midrule
A3 & Nearest & Native Detect & 53.006 & 40.972 & 42.431 & 25.542 & 9.763 & 29.166 \\
A4 & DySample & Native Detect & \textbf{53.747} & 40.525 & 42.388 & 25.569 & 9.773 & 29.182 \\
A5-DyOff & Nearest & OS-CVTIA & 52.233 & 41.155 & 42.256 & 25.624 & 10.032 & 29.864 \\
\textbf{A5} & \textbf{DySample} & \textbf{OS-CVTIA} & 53.367 & \textbf{41.572} & \textbf{42.737} & \textbf{25.713} & 10.042 & 29.879 \\
\bottomrule
\end{tabular}
\end{adjustbox}
\end{table*}

With the native detection head, replacing nearest-neighbor upsampling with DySample changes precision, recall, mAP$_{50}$, and mAP$_{50:95}$ by $+0.741$, $-0.447$, $-0.043$, and $+0.027$ percentage points, respectively. Thus, DySample alone produces only a marginal net AP change in this setting. Under OS-CVTIA, however, DySample improves precision, recall, mAP$_{50}$, and mAP$_{50:95}$ by 1.134, 0.417, 0.481, and 0.089 percentage points, respectively, relative to A5-DyOff. The results therefore support a complementary relationship between content-dependent spatial alignment and the subsequent task-interaction head, rather than a claim that DySample independently produces a large gain in every configuration.

\subsubsection{Ablation of OS-CVTIA}
\label{subsubsec:oscvtia_ablation}
Table~\ref{tab:oscvtia_ablation} compares OS-CVTIA with the native detection head and an alternative H2 interaction configuration. In the final H1 design, boundary-sensitive and context-sensitive macro--micro representations are used to modulate the localization and classification pathways. The classification interaction includes a small relaxation term before multiplicative modulation, and independent learnable scaling factors regulate the classification and localization modulation intensities.

\begin{table*}[!t]
\centering
\caption{Ablation of OS-CVTIA interaction designs on VisDrone2019.}
\label{tab:oscvtia_ablation}
\renewcommand{\arraystretch}{1.08}
\setlength{\tabcolsep}{5.0pt}
\footnotesize
\begin{tabular}{lccccc}
\toprule
Detection head & Interaction design & P (\%) & R (\%) & mAP$_{50}$ (\%) & mAP$_{50:95}$ (\%) \\
\midrule
Native Detect & -- & \textbf{53.747} & 40.525 & 42.388 & 25.569 \\
OS-CVTIA-H2 & Weakened micro interaction & 52.032 & \textbf{41.706} & 42.603 & 25.533 \\
\textbf{OS-CVTIA-H1} & \textbf{Full interaction + relaxation} & 53.367 & 41.572 & \textbf{42.737} & \textbf{25.713} \\
\bottomrule
\end{tabular}
\end{table*}

Compared with the native detection head, OS-CVTIA-H1 changes precision, recall, mAP$_{50}$, and mAP$_{50:95}$ by $-0.380$, $+1.047$, $+0.349$, and $+0.144$ percentage points, respectively. This pattern suggests that macro--micro task interaction helps retain weak object responses and yields a more favorable overall precision--recall balance, although it does not improve every detection metric.

H1 also improves precision, mAP$_{50}$, and mAP$_{50:95}$ over H2 by 1.335, 0.134, and 0.180 percentage points, respectively, while H2 has a 0.134-point recall advantage. These results motivate the use of H1 as the final configuration because it provides the highest overall AP among the evaluated head designs. The H1--H2 comparison is reported without separate complexity values because the two variants share the same static structural complexity; earlier profiling differences were caused by inconsistent fused/unfused counting conventions rather than by a meaningful architectural difference.

With the architectural configuration fixed to the final A5 design, we next examine whether the localization objective should retain the native CIoU formulation or incorporate an NWD-based term for tiny-object regression.

\subsubsection{Localization-Loss Selection}
\label{subsubsec:loss_ablation}

Because NWD was proposed to improve localization sensitivity for tiny objects~\citep{wang2021nwd}, we further examine whether it provides an additional benefit to the final HDMamba-YOLO-B detector. The architecture is fixed to the complete A5 configuration, and the dataset split, optimizer, augmentation policy, random seed, training schedule, and all remaining loss terms are kept unchanged. Only the relative contribution of CIoU and NWD in the localization objective is varied. Three settings are evaluated: pure CIoU, an equal-weight CIoU--NWD mixture, and pure NWD.

\begin{table*}[!t]
\centering
\caption{Controlled ablation of the localization objective on VisDrone2019. The HDMamba-YOLO-B architecture and all remaining training settings are kept unchanged; only the CIoU/NWD weighting is varied.}
\label{tab:loss_ablation}
\renewcommand{\arraystretch}{1.08}
\setlength{\tabcolsep}{5.0pt}
\footnotesize
\begin{adjustbox}{max width=\textwidth}
\begin{tabular}{lcccccc}
\toprule
Localization objective &
$w_{\mathrm{CIoU}}$ &
$w_{\mathrm{NWD}}$ &
P (\%) &
R (\%) &
mAP$_{50}$ (\%) &
mAP$_{50:95}$ (\%) \\
\midrule
\textbf{Pure CIoU} &
\textbf{1.0} &
\textbf{0.0} &
\textbf{53.367} &
\textbf{41.572} &
\textbf{42.737} &
\textbf{25.713} \\

CIoU + NWD &
0.5 &
0.5 &
51.909 &
40.553 &
41.691 &
25.011 \\

Pure NWD &
0.0 &
1.0 &
52.819 &
39.523 &
40.584 &
23.898 \\
\bottomrule
\end{tabular}
\end{adjustbox}
\end{table*}

As shown in Table~\ref{tab:loss_ablation}, pure CIoU provides the best result across all four detection metrics. Compared with the equal-weight CIoU--NWD formulation, pure CIoU improves precision, recall, mAP$_{50}$, and mAP$_{50:95}$ by 1.458, 1.019, 1.046, and 0.702 percentage points, respectively. Replacing CIoU entirely with NWD leads to a larger degradation in AP, with mAP$_{50}$ decreasing from 42.737\% to 40.584\% and mAP$_{50:95}$ from 25.713\% to 23.898\%.

The results show that introducing NWD does not provide an additional localization benefit for the final HDMamba-YOLO-B configuration under the present VisDrone training protocol. We therefore retain pure CIoU as the localization objective for all final HDMamba-YOLO results. With both the architectural configuration and localization objective determined, we next examine the optimization behavior of the final model.

\subsubsection{Training Convergence}
\label{subsubsec:convergence}
\begin{figure*}[!t]
    \centering
    \includegraphics[
        width=0.94\textwidth,
        keepaspectratio
    ]{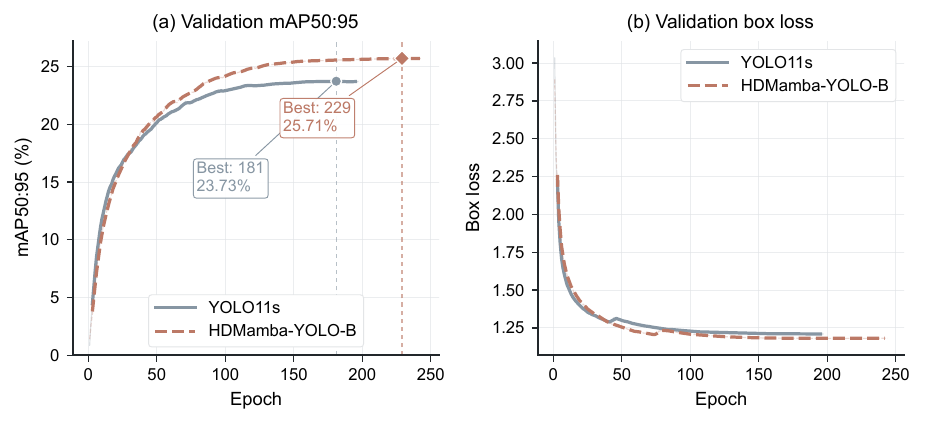}
    \caption{Training convergence on VisDrone2019 for YOLO11s and HDMamba-YOLO-B. The two panels report validation mAP$_{50:95}$ and validation box loss, respectively. Both models exhibit stable optimization, while HDMamba-YOLO-B converges to a higher validation accuracy and a lower box-loss level. The figure is intended to compare final optimization behavior rather than convergence speed.}
    \label{fig:visdrone_convergence}
\end{figure*}

Figure~\ref{fig:visdrone_convergence} compares the validation trajectories of YOLO11s and HDMamba-YOLO-B under the same VisDrone training protocol. Both models converge stably under early stopping, but the proposed model reaches a higher validation mAP$_{50:95}$ and maintains a lower validation box-loss trajectory near convergence. The best recorded mAP$_{50:95}$ values are 23.73\% for YOLO11s and 25.71\% for HDMamba-YOLO-B, reached at epochs 181 and 229, respectively. Over the final 20 recorded epochs, the corresponding mAP$_{50:95}$ means are 23.708\% and 25.701\%, with standard deviations of 0.017 and 0.006 percentage points. These curves support stable late-stage optimization and a higher converged accuracy for HDMamba-YOLO-B; because the best epoch occurs later for the proposed model, they should not be interpreted as evidence of faster convergence.

\subsection{Discussion}
\label{subsec:discussion}
The experiments provide a consistent but nuanced picture of HDMamba-YOLO. On VisDrone2019, the Tiny, Lite, and Base variants span three parameter ranges and exhibit the expected accuracy increase as model capacity grows. The Base variant achieves the strongest performance in the compact comparison group, while the Lite and Tiny variants provide lower-complexity operating points. On AI-TOD, HDMamba-YOLO-B improves the principal AP metrics over the similarly sized YOLO11s and remains reasonably close to the substantially larger YOLOv8m on several tiny- and small-object metrics. At the same time, the lower AP$_{vt}$ and AR$_{vt}$ values show that objects represented by only 2--8 pixels remain a challenging regime for the current architecture.

The controlled ablations support the stage-wise design rationale. EVSS provides the largest gain when introduced into the perception stage, whereas the SSM-dominant neck control performs slightly below the final convolutional reconstruction design. The CNN Full-Control result provides the complementary observation that replacing the EVSS perception stages with CNN/InRes processing substantially reduces AP under the same downstream configuration. Together, these experiments support a functional allocation in which state-space modeling is used for contextual perception and convolutional operators are retained for local two-dimensional reconstruction in the neck.

The reconstruction analysis further shows that the Local and Context branches of C3k2-ASSAF exhibit different response tendencies. The Local-only variant favors precision and stricter localization, whereas the Context-only variant favors recall. Their combination provides the highest mAP$_{50}$ and the best overall precision--recall balance while maintaining essentially the same mAP$_{50:95}$ as Local-only. This result supports the local--context dual-path formulation while also motivating a restrained interpretation of the adaptive gate: the gate is retained as a lightweight modulation mechanism, but the principal evidence concerns branch complementarity rather than a large isolated gate contribution.

The alignment and interaction studies reveal a similar complementary relationship. DySample produces only a marginal AP change with the native detection head, but its benefit becomes more consistent when paired with OS-CVTIA. OS-CVTIA, in turn, primarily improves recall and overall AP rather than precision alone. These results are consistent with the intended processing sequence of reconstruction, cross-scale alignment, and task interaction, while avoiding the stronger claim that each module independently improves every metric.

Finally, the stage-transition study shows that the final PhasePatchMerging2D configuration provides higher AP than both EfficientDownsample and LightweightPhaseDownsample at a modest complexity increase. Because LightweightPhaseDownsample also uses four spatial phases, the result suggests that the gain is associated with the complete phase-aware merge and projection formulation rather than with phase sampling alone. The localization-loss ablation further shows that pure CIoU provides the best performance among the three evaluated objectives for the final HDMamba-YOLO-B configuration.

Overall, the results support the central design of HDMamba-YOLO as a coordinated perception--reconstruction--alignment--interaction pipeline. The evidence is strongest when interpreted within the controlled UAV small-object setting used in this work: contextual state-space processing is effective in the backbone, local convolutional reconstruction remains useful in the neck, dynamic alignment and task interaction provide complementary refinements, and the resulting architecture offers multiple accuracy--complexity operating points for aerial small-object detection.

\section{Conclusion}

In this work, we proposed Hybrid Dual-domain Mamba-YOLO (HDMamba-YOLO), an efficient UAV small-object detector built on a stage-wise heterogeneous SSM--CNN architecture. Rather than applying a single operator family uniformly throughout the detector, HDMamba-YOLO organizes feature processing according to four successive functional requirements: perception, reconstruction, alignment, and interaction. EfficientVMamba-based EVSS is employed in the relatively high-resolution backbone stages to establish long-range contextual perception, while PhasePatchMerging2D performs phase-aware hierarchical transitions before deeper semantic abstraction. CNN-style processing is retained in the neck to provide explicit local two-dimensional reconstruction, thereby complementing the contextual representation learned by the state-space backbone.

To support this stage-wise allocation, DST-Wrapper re-encodes the deepest backbone representation at the perception--reconstruction interface, and Native C3k2-ASSAF introduces repeated local-detail and bounded-context reconstruction while preserving the native split--iterative--concatenate feature-reuse topology of C3k2. DySample further performs content-adaptive cross-scale resampling at the top-down fusion stages to improve spatial correspondence between features of different resolutions. Finally, OS-CVTIA introduces macro--micro receptive-field interaction and asymmetric task modulation for localization and classification before the native YOLO prediction branches. Together, these components instantiate the proposed perception--reconstruction--alignment--interaction (P--R--A--I) pipeline.

Experiments on VisDrone2019 and AI-TOD demonstrate that HDMamba-YOLO provides a competitive accuracy--complexity trade-off for UAV small-object detection in dense aerial scenes. On VisDrone2019, the canonical HDMamba-YOLO-B achieves 42.737\% mAP$_{50}$ and 25.713\% mAP$_{50:95}$ with 10.042M parameters and 29.879 GFLOPs. The reduced-capacity HDMamba-YOLO-Tiny and HDMamba-YOLO-Lite variants further provide lower-complexity operating points while retaining the same stage-wise design principle. The progressive ablation studies show that the individual components contribute complementary improvements, while the controlled SSM--CNN architecture comparison indicates that, under the present experimental setting, allocating state-space modeling to contextual perception and CNN-style processing to local reconstruction is more effective than extending state-space aggregation uniformly into the neck. Qualitative analyses further reveal more concentrated feature responses, the recovery of additional instances in selected dense scenes, and complementary local--context response patterns within ASSAF.

The results also reveal remaining challenges. In particular, extremely tiny targets remain difficult under the AI-TOD very-tiny regime, where weak appearance evidence and severe spatial quantization continue to limit recall. Future work will therefore investigate stronger localization-aware learning objectives and feature alignment strategies for very-tiny targets, as well as more efficient variants of the current architecture for real-time edge deployment. The stage-wise global--local perception framework may also be extended to broader low-altitude visual perception tasks, including traffic monitoring, emergency response, and other resource-constrained aerial applications.

\bibliographystyle{elsarticle-harv}
\bibliography{cas-refs}

\end{document}